# Spacecraft design optimisation for demise and survivability

Mirko Trisolini[a]*, Hugh G. Lewis[a], Camilla Colombo[b]

[a] *Astronautics Research Group, University of Southampton, Southampton , SO17 1BJ, United Kingdom,*
*m.trisolini@soton.ac.uk, H.G.Lewis@soton.ac.uk*
[b] *Department of Aerospace Science and Technology, Politecnico di Milano, Via La Masa 34, 20133, Milan, Italy,*
*camilla.colombo@polimi.it*
* Corresponding Author

**Abstract**
Among the mitigation measures introduced to cope with the space debris issue there is the de-orbiting of decommissioned satellites. Guidelines for re-entering objects call for a ground casualty risk no higher than $10^{-4}$. To comply with this requirement, satellites can be designed through a design-for-demise philosophy. Still, a spacecraft designed to demise through the atmosphere has to survive the debris-populated space environment for many years. The demisability and the survivability of a satellite can both be influenced by a set of common design choices such as the material selection, the geometry definition, and the position of the components inside the spacecraft. Within this context, two models have been developed to analyse the demise and the survivability of satellites. Given the competing nature of the demisability and the survivability requirements, a multi-objective optimisation framework was developed, with the aim to identify trade-off solutions for the preliminary design of satellites. As the problem is nonlinear and involves the combination of continuous and discrete variables, classical derivative based approaches are unsuited and a genetic algorithm was selected instead. The genetic algorithm uses the developed demisability and survivability criteria as the fitness functions of the multi-objective algorithm. The paper presents a test case, which considers the preliminary optimisation of tanks in terms of material, geometry, location, and number of tanks for a representative Earth observation mission. The configuration of the external structure of the spacecraft is fixed. Tanks were selected because they are sensitive to both design requirements: they represent critical components in the demise process and impact damage can cause the loss of the mission because of leaking and ruptures. The results present the possible trade off solutions, constituting the Pareto front obtained from the multi-objective optimisation.

**Keywords:** design-for-demise, survivability, multi-objective optimisation, tanks

**Nomenclature**

| | | |
|---|---|---|
| $a$ | Semi-major axis | Km |
| $A$ | Area | m² |
| $C$ | Speed of sound | m/s |
| $C_D$ | Drag coefficient | |
| $CF$ | Correction factor for component mutual shielding | |
| $C_m$ | Heat capacity | J/Kg-K |
| $D$ | Diameter | m |
| $d$ | Lateral size of a component in the impact plane | m |
| $E_0$ | Maximum allowed displacement from the nominal ground track at the equator | Km |
| $\bar{F}_q$ | Motion and shape averaged shape factor for heat flux predictions | |
| $G$ | Universal gravitational constant | |
| $g_\varphi$ | Polar component of the gravitational acceleration | m/s² |
| $g_0$ | Gravitational acceleration at sea level | m/s² |
| $g_R$ | Radial component of the gravitational acceleration | m/s² |
| $h$ | Altitude | Km |
| $h_f$ | Heat of fusion | J/kg |
| $Isp$ | Specific impulse | s |

| Symbol | Description | Units |
|---|---|---|
| $K1$ | Factor to account for the additional tank volume for the pressuring gas | |
| $K2$ | Factor to account for the separation between two tanks | |
| $l$ | Distance between two tanks | m |
| $L$ | Side length of the spacecraft | m |
| $L$ | Length | m |
| $m$ | Mass | Kg |
| $M_E$ | Earth's mass, $5.97 \times 10^{24}$ | Kg |
| $N$ | Total number of spacecraft components | |
| $n_t$ | Number of tanks | |
| $Pp$ | Penetration probability | |
| $r$ | Radius | m |
| $R_E$ | Earth's radius, 6371800 | m |
| $r_n$ | Nose radius | m |
| $S$ | Cross-section of the spacecraft | m$^2$ |
| $s$ | Stand-off distance | m |
| $SF$ | Tank pressure safety factor | |
| $t$ | Thickness | m |
| $t_m$ | Mission duration | years |
| $T_m$ | Melting temperature | K |
| $V$ | Velocity | m/s |
| $v$ | Volume | m$^3$ |
| $\alpha$ | Cone ejecta spread angle | ° |
| $\gamma$ | Flight path angle | ° |
| $\varepsilon$ | Emissivity | |
| $\theta$ | Impact angle | ° |
| $\lambda$ | Longitude | ° |
| $\rho$ | Density | Kg/m$^3$ |
| $\sigma$ | Stefan-Boltzmann constant, $5.67 \times 10^{-8}$ | W/m$^2$/K$^4$ |
| $\sigma_u$ | Ultimate tensile strength | MPa |
| $\sigma_y$ | Yield strength | MPa |
| $\varphi$ | Latitude | ° |
| $\phi$ | Debris flux | 1/m$^2$/yr |
| $\chi$ | Heading angle | ° |
| $\omega$ | Angular velocity | °/s |

Subscripts

| | | |
|---|---|---|
| $0$ | Nominal orbit | |
| $1$ | Start orbit for the Hohmann transfer | |
| $2$ | Final orbit of the Hohmann transfer | |
| $atm$ | Atmosphere | |
| $BLE$ | Relative to the probability of penetrating an internal component | |
| $C$ | Relative to the critical diameter | |
| $comp$ | Relative to the probability of impacting a component after a first impact on the vulnerable zone | |
| $decay$ | Relative to decaying correction manoeuvres | |
| $disp$ | Relative to disposal manoeuvres | |
| $e$ | Earth | |
| $ejecta$ | Relative to the debris cone produced after impact | |
| $f$ | Fuel | |
| $fin$ | Final condition | |

| | |
|---|---|
| *in* | Initial condition |
| *inc* | Relative to inclination change manoeuvres |
| *inj* | Relative to orbit injection errors |
| *mat* | Material |
| *p* | Debris particle |
| *s* | Spacecraft |
| *sec* | Relative to secular variations of the orbital parameters |
| *struct* | Relative to the impact on the external structure inside the vulnerable zone |
| *t* | Tank |
| *target* | Target component of the impact probability analysis |
| *tot* | Total |
| *VZ* | Relative to the vulnerable zone |
| *w* | Wall |

**Abbreviation**

| | |
|---|---|
| DAS | Debris Assessment Software |
| LMF | Liquid Mass Fraction |
| BLE | Ballistic Limit Equation |
| SRL | Schafer-Ryan-Lambert |
| NSGA | Non-dominated Sorting Genetic Algorithm |
| PNP | Probability of no-penetration |

## 1 Introduction

In the past two decades, the attention towards a more sustainable use of outer space has increased steadily. The major space-faring nations and international committees have proposed a series of debris mitigation measures [1, 2] to protect the space environment. Among these mitigation measures, the de-orbiting of spacecraft at the end of their operational life is recommended in order to reduce the risk of collisions in orbit.

However, re-entering spacecraft can pose a risk to people and property on the ground. Consequently, the re-entry of disposed spacecraft needs to be analysed and its compliancy with international regulations has to be assessed. In particular, the casualty risk for people on the ground related to the re-entry of a spacecraft needs to be below the limit of $10^{-4}$ if an uncontrolled re-entry strategy is to be adopted [3, 4]. A possible strategy to limit the ground casualty risk is to use a design-for-demise philosophy, where most (if not all) of the spacecraft will not survive the re-entry process. The implementation of design for demise strategies [5-7] may favour the selection of uncontrolled re-entry disposal options over controlled ones, leading to a simpler and cheaper alternative for the disposal of a satellite at the end of its operational life [6, 7]. However, a spacecraft designed for demise still has to survive the space environment for many years. As a large number of space debris and meteoroids populates the space around the Earth, a spacecraft can suffer impacts from these particles, which can be extremely dangerous, damaging the spacecraft or even causing the complete loss of the mission [8-10]. This means that the spacecraft design has also to comply with the requirements arising from the survivability against debris impacts.

The demisability and survivability of a spacecraft are both influenced by a set of common design choices, such as the material of the structure, its shape, dimension and position inside the spacecraft. It is important to consider such design choices and how they influence the mission's survivability and demisability from the early stages of the mission design process [7]. In fact, taking into account these requirements at a later stage of the mission may cause an inadequate integration of these design solutions, leading to a delayed deployment of the mission and to an increased cost of the project. On the other hand, an early consideration of such requirements can favour cheaper options such as the uncontrolled re-entry of the satellite, whilst maintaining the necessary survivability and, thus, the mission reliability.

With these considerations, two models have been developed [11] to assess the demisability and the survivability of simplified mission designs as a function of different design parameters. Two criteria are presented to evaluate the degree of demisability and survivability of a spacecraft configuration. Such an analysis can be carried out on many different kinds of missions, provided that they can be disposed through atmospheric re-entry and they experience impacts from debris particles during their operational life. These characteristics are common to a variety of missions;

however, it was decided to focus the current analysis on Earth observation and remote sensing missions. Many of these missions exploit sun-synchronous orbits due to their favourable characteristics, where a spacecraft passes over any given point of the Earth's surface at the same local solar time. Because of their appealing features, sun-synchronous orbits have high commercial value. Alongside their value from the commercial standpoint, they are also interesting for a combined survivability and demisability analysis. Sun-synchronous missions can in fact be disposed through atmospheric re-entry. They are also subject to very high debris fluxes [12] making them a perfect candidate for the purpose of this study.

Given the competing nature of the demisability and survivability requirements, a multi-objective optimisation framework has been developed, with the aim to find trade-off solutions for the preliminary design of satellites. As the problem is nonlinear and involves the combination of continuous and discrete variables, classical derivative based approaches are unsuited and a genetic algorithm was selected instead. The genetic algorithm uses the previously described demise and survivability criteria as the fitness functions of the multi-objective algorithm.

The paper presents a test case, which considers the preliminary optimisation of tanks in terms of material, geometry, location, and number of tanks for representative sun-synchronous missions. The configuration of the external structure of the spacecraft was fixed. Tanks were selected because they are interesting for both the survivability and the demisability. They represent critical components in the demisability analysis as they usually survive the atmospheric re-entry. They are also components that need particular protection from the impact against space debris because the impact with debris particles can cause leaking or ruptures, which can compromise the mission success. Different configurations were analysed as a function of the characteristics of the tank assembly and of the mission itself, such as the mission duration and the mass class of the spacecraft. The results are presented in the form of Pareto fronts, which represent the different possible trade-off solutions.

**2   Demisability and Survivability Models**

In order to carry out a combined demisability and survivability analysis of a spacecraft configuration it was necessary to develop two models [11, 13-15]. One model allows the analysis of the atmospheric re-entry of a simplified spacecraft configuration in order to evaluate its demisability. The other model carries out a debris impact analysis and returns the penetration probability of the satellite as a measure of its survivability. As these two models need to be implemented into an optimisation framework, much effort was made to maintain a comparable level of detail and computational time between them.

*2.1  Demisability model*

The demisability model consists of an object-oriented code [16-18]. The main features of this type of code is the fast simulation of the re-entry of a spacecraft that is schematised using primitive shapes such as spheres, cubes, cylinders, and flat plates. These primitive shapes are used as a simplified representation of both the main spacecraft structure and internal components. The different parts of the spacecraft are defined in a hierarchical fashion. The first level is constituted by the main spacecraft structure (commonly referred as the parent object). Here the overall spacecraft mass and dimensions are defined, as well as the solar panels area. The second level defines the external panels of the main structure. This gives the opportunity to the user to specify different characteristics for the external panels, such as the material, the thickness, and the type of panel (e.g. honeycomb sandwich panel). The third level defines the main internal components such as tanks and reaction wheels. An additional level can also be used for the definition of sub-components like battery cells.

The main structure of the code is represented in Figure 1. First, the software requires a set of inputs from the user, which has to provide the initial conditions of the re-entry trajectory in the form of longitude, latitude, altitude, relative velocity, flight path angle, and heading angle. The second required input from the user is the spacecraft configuration. This is a file describing the characteristics of each component in the configuration. An example of the structure of such file is represented in Table 1. Additionally, the user can specify the integration options, such as the time step, and the relative and absolute tolerances.

Table 1: Example of spacecraft configuration to be provided to the software.

| ID | Name | Parent | Shape | Mass (kg) | Length (m) | Radius (m) | Width (m) | Height (m) | Quantity |
|---|---|---|---|---|---|---|---|---|---|
| 0 | Spacecraft | n/a | Box | 2000 | 3.5 | n/a | 1.5 | 1.5 | 1 |
| 1 | Tank | 0 | Sphere | 15 | n/a | 0.55 | n/a | n/a | 1 |
| 2 | BattBox | 0 | Box | 5 | 0.6 | n/a | 0.5 | 0.4 | 1 |
| 3 | BattCell | 2 | Box | 1 | 0.1 | n/a | 0.05 | 0.05 | 20 |
| … | … | … | … | … | … | … | … | … | … |

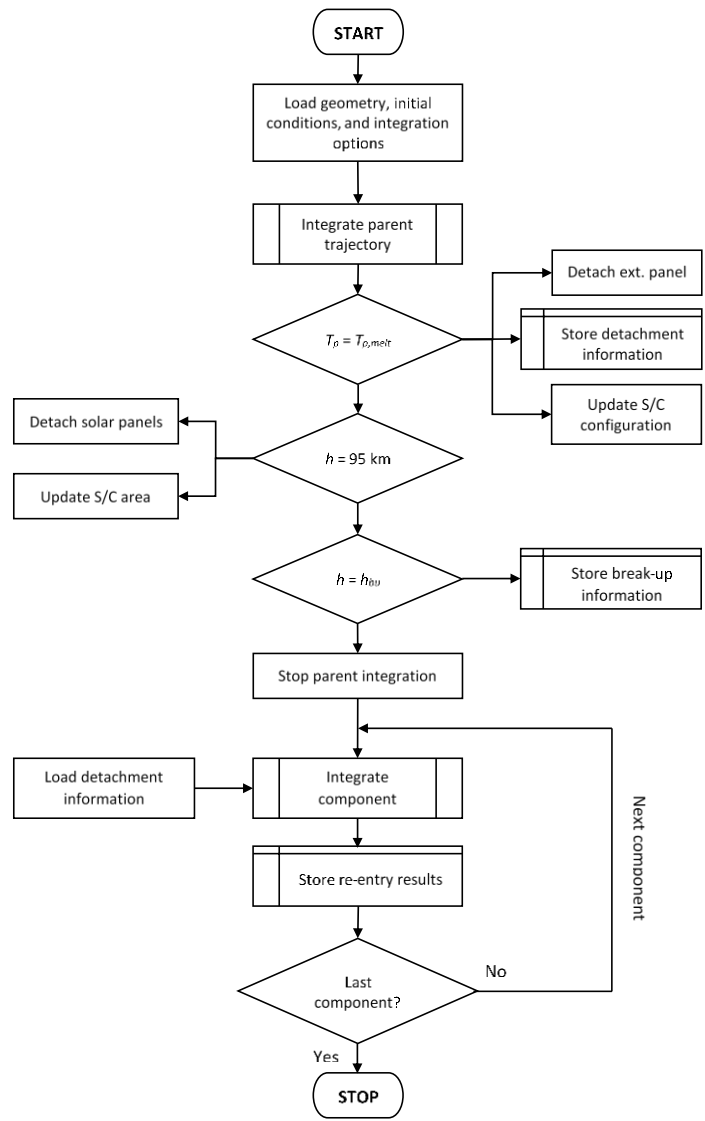

Figure 1: Flow diagram of the main structure of the re-entry software.

After the initialisation is complete, the trajectory of the parent object is simulated. This first part of the simulation carries on until the main break-up altitude is reached and only the parent structure can interact with the external heat flux. The internal components do not experience any heat load during this phase. The break-up altitude is user-defined but the default value is set to 78 km, which is the standard value used in most destructive re-entry software [19]. During this first phase of the simulation, the software can take into account the occurrence of some specific events, such as the detachment of the solar panels or the external panels of the main structure. It is considered that the solar panels separate from the main structure at a fixed altitude, equal to 95 km [20, 21]. When this altitude is reached, the solar panels are simply removed from the simulation, with the consequent change in the aerodynamics of the structure, as the area of the solar panels is not considered anymore in the contribution to the average cross-section of the spacecraft. The re-entry of the solar panels after detachment is not simulated, as they are considered to always demise. The detachment of the external panels of the structure is instead triggered by the temperature of the panels themselves. Once the temperature reaches the melting the temperature of the panel material, the panel is considered to detach [22]. If an internal component is attached to the panel, also the component is considered to detach from the main structure. At this point, the detachment conditions for each detached panel and component are stored for a later use and the mass of the spacecraft is updated accordingly. The first part of the simulation ends with the main spacecraft reaching the break-up altitude. At this point, the break-up state is also stored, as it will be used as the initial state for all the internal components released at break-up. After the break-up event, the parent spacecraft is removed and the re-entry of the internal components and of the external panels is simulated. To each component is associated an initial condition that is the previously stored detachment conditions.

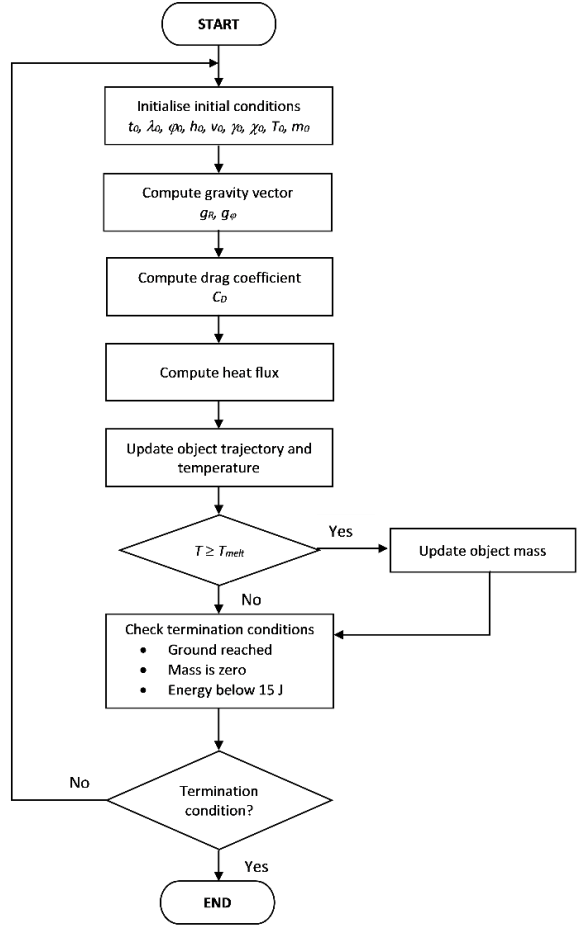

Figure 2: Flow diagram for the procedure used in the integration of the re-entry trajectory.

When the re-entry of the component has to be simulated, its initial condition is loaded. The simulation continues until the component reaches the ground, or it is completely demised, or its energy is below the threshold of 15 J [3]. When one of these events occurs, the simulation is stopped and the re-entry output is stored. We store the final mass, cross-section, landing location and impact energy of each surviving object, and the demise altitude of each demised object. The simulation procedure is repeated for each internal component and external panel. The procedure followed for the integration of the trajectory of the main spacecraft and the internal components is summarised in Figure 2. The trajectory is simulated with a three degree-of-freedom ballistic dynamics, where the computation of the attitude motion of the objects is neglected and the attitude motion is predefined and assumed random tumbling. The Earth's atmosphere is modelled using the 1976 U.S Standard Atmosphere [23]. The Earth's gravitational acceleration is modelled using a zonal harmonic gravity model up to degree four [24]. The radial and polar contributions to the gravity vector can be expressed as

$$g_R = -\frac{GM_E}{r}\left\{1 - \frac{3}{2}J_2\left(\frac{R_E}{r}\right)^2[3\sin^2\varphi - 1] - 2J_3\left(\frac{R_E}{r}\right)^3[5\sin^3\varphi - 3\sin\varphi] - \frac{5}{8}J_4\left(\frac{R_E}{r}\right)^4[35\sin^4\varphi - 30\sin^2\varphi + 3]\right\} \quad (1)$$

$$g_\phi = \frac{3GM_E}{r^2}\left(\frac{R_E}{r}\right)^2 \sin\varphi\cos\varphi\left\{J_2 + \frac{1}{2}J_3\left(\frac{R_E}{r}\right)[5\sin^2\varphi - 1]\right.$$
$$\left. + \frac{5}{6}J_4\left(\frac{R_E}{r}\right)^2[7\sin^2\varphi - 1]\right\} \quad (2)$$

where $g_R$ and $g_\phi$ are the radial and polar components of the gravitational acceleration. $G$ is the universal gravitational constant, $M_E$ is the mass of the Earth, $R_E$ is the equatorial radius of the Earth, $r$ is the distance between the satellite and the centre of the Earth, $\varphi$ is the latitude. Finally, the $J_k s$ ($k = 2, 3, 4$) are the zonal harmonic coefficients, which take into account the effect of the Earth's oblateness and not uniform mass distribution.

Alongside the gravitational forces, the aerodynamic forces are needed in order to predict the trajectory of the spacecraft and its components. The aerodynamic force acting on the object during the re-entry can be expressed as

$$F_D = \frac{1}{2}\rho V^2 \cdot S \cdot C_D \quad (3)$$

where $\rho$ is the air density, $V$ is the air velocity, $S$ is the cross section and $C_D$ is the drag coefficient. The drag coefficient is computed using averages over the motion and shape of the object [25-27]. In addition, drag coefficients for both free-molecular and continuum regimes are taken into account. During its descent, in fact, the spacecraft encounters very rarefied air at high altitude such that the continuum flow model cannot be adopted and a free-molecular approximation needs to be used. The free-molecular and continuum drag coefficient for each elementary shape are summarised in Table 2.

Table 2: Free-molecular and continuum drag coefficients.

| Shape | Free-molecular $C_D$ | Continuum $C_D$ | Reference area |
|---|---|---|---|
| Sphere | 2.0 | 0.92 | $\pi \cdot D^2 / 4$ |
| Box | $1.0 \cdot (A_x + A_y + A_z)/A_y$ | $0.46 \cdot (A_x + A_y + A_z)/A_y$ | $A_y$ |
| Cylinder | $1.57 + 0.785 \cdot D/L$ | $0.7198 + 0.326 \cdot D/L$ | $D \cdot L$ |
| Flat Plate | 1.03 | 0.46 | $A$ |

where $D$ is the diameter of the sphere or the cylinder, $L$ is the length of the cylinder, and $A_x$, $A_y$, and $A_z$ are the areas of the sides of the box. The areas are ordered such that $A_x \geq A_y \geq A_z$.

Similarly to the aerodynamics, even the aerothermodynamics is taken into account using motion and shape averaged coefficients. In general, the heat flux on an object is computed as follows

$$\dot{q}_{av} = \bar{F}_q \cdot \dot{q}_{ref} \quad (4)$$

Where $\dot{q}_{av}$ is the average heat load on the object in W/m², $\bar{F}_q$ is the averaging factor depending on the shape of the object, its motion, and the re-entry regime (i.e. free-molecular or continuum), and $\dot{q}_{ref}$ is the reference heat load on the object. The reference heat load is different from the free-molecular and the continuum case. In the first case, is represented by the heat rate on a flat plate perpendicular to the free-stream flow that is

$$\dot{q}_{rem}^{fm} = 11356.6 \cdot \left(\frac{a \cdot \rho_0 \cdot V_0^3}{1556}\right) \quad (5)$$

where $a$ is the thermal accommodation coefficient. For metallic material the accommodation coefficient is usually around 0.9, which was adopted as the constant in the current analysis [27]. $\rho_0$ is the free-stream density and $V_0$ is the free-stream velocity. In the continuum case, instead, the reference heat load is the heat flux at the stagnation point of a sphere, which is the Detra, Kemp and Riddel (DKR) correlation [27, 28]

$$\dot{q}_{ref}^{cont} = 1.99876 \times 10^8 \cdot \sqrt{\frac{0.3048}{r_n}} \cdot \sqrt{\frac{\rho_0}{\rho_{SL}}} \cdot \left(\frac{V_0}{7924.8}\right)^{3.15} \cdot \frac{h_s - h_w}{h_s - h_{w_{300}}} \quad (6)$$

where $r_n$ is the curvature radius at stagnation point, and $\rho_{SL}$ is the air density at sea-level. $h_s$ is the stagnation point enthalpy, $h_w$ is the wall enthalpy and $h_{w300}$ is the wall enthalpy at 300 K. The averaging factors for the computation of the heat flux are summarised in Table 3.

Table 3: Free-molecular and continuum averaging factors for the heat load computation.

| Shape | Free-molecular $\bar{F}_q$ | Continuum $\bar{F}_q$ | Reference area |
|---|---|---|---|
| Sphere | 0.255 | 0.345 | $\pi \cdot D^2$ |
| Box | $\dfrac{\pi D_{eq} L_{eq} \begin{pmatrix} 0.1275 \cdot D_{eq}/L_{eq} + 0.785 \cdot Y \\ + 0.5 \cdot Z \end{pmatrix}}{A_{wet}^{box}}$ | $\dfrac{\pi D_{eq} L_{eq} \begin{pmatrix} 0.179 + 0.1615 \cdot D_{eq}/L_{eq} \\ + 0.333 \cdot B \end{pmatrix}}{A_{wet}^{box}}$ | $A_{wet}^{box}$ |
| Cylinder | $\dfrac{0.255 \cdot D/L + 1.57 \cdot Y + Z}{2 + D/L}$ | $\dfrac{0.358 + 0.323 \cdot D/L + 0.666 \cdot B}{2 + D/L}$ | $\pi D \left( \dfrac{D}{2} + L \right)$ |
| Flat Plate | 0.255 | $0.233 \cdot \dfrac{A_{wet}^{d,eq}}{A_{wet}^{fp}}$ | $A_{wet}^{fp}$ |

To compute the box shape factors an equivalent cylinder has been used [27]. $D_{eq} = 2 \cdot \sqrt{W \cdot H}$ is the equivalent diameter and $L_{eq} = L$ is the equivalent length, with $L$, $W$, and $H$ being the length, width and height of the box. $A_{wet}^{box} = 2 \cdot (LW + LH + WH)$ Awet is the external surface of the box, and $Y$, $Z$, and $B$ are values dependent on the Mach number and can be extracted from plots available at [25]. $A_{wet}^{fp}$ is the external area of the flat plate, and $A_{wet}^{d,eq}$ is the external area of an equivalent disk with an equivalent diameter of $D_{eq} = W/2$.

The trajectory is described by a set of three degree-of-freedom equations of motion, where the spacecraft is treated as a material point and its attitude motion is established a priori. As previously described, the aerodynamics and aerothermodynamics are thus expressed using motion and shape averaged coefficients. As the aerodynamic forces are due to the motion of the vehicle relative to the atmosphere of the planet, a reference frame fixed with the atmosphere is required to determine the aerodynamics loads on the vehicle. Since a planet's atmosphere rotates with it, it is possible to define a planet-fixed reference frame in order to express the equations of the atmospheric flight (Figure 3).

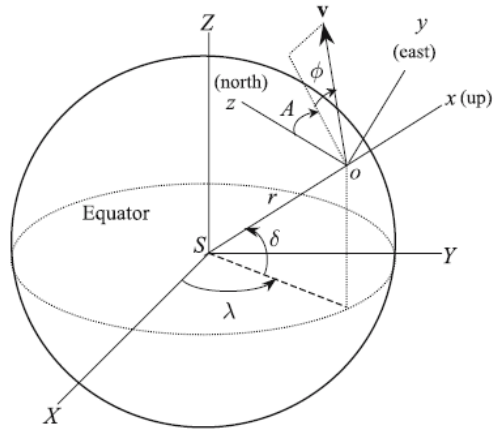

Figure 3: Planet fixed (SXYZ) and local horizon (oxyz) frames [24].

The complete set of equations of motion is summarised in Eqs. (7).

$$\dot{r} = V \sin \gamma$$

$$\dot{\varphi} = \frac{V}{r} \cos \gamma \cos \chi$$

$$\dot{\lambda} = \frac{V \cos \gamma \sin \chi}{r \cos \varphi}$$

$$\dot{V} = -F_D/m + g_R \sin \gamma - g_\phi \cos \gamma \cos \chi - \omega^2 r \cos \varphi \left( \cos \gamma \cos \chi \sin \varphi - \sin \gamma \cos \varphi \right)$$

$$\dot{\chi} = \frac{V}{r} \cos \gamma \sin \chi \tan \varphi + \frac{g_\varphi}{V} \frac{\sin \chi}{\cos \gamma} + \frac{\omega^2 r}{V} \frac{\sin \chi \sin \varphi \cos \varphi}{\cos \gamma} - \frac{2\omega}{\cos \gamma} \left( \tan \gamma \cos \chi \cos \varphi - \sin \varphi \right)$$

$$\dot{\gamma} = \frac{V}{r} \cos \gamma + \frac{g_R}{V} \cos \gamma + \frac{g_\phi}{V} \sin \gamma \cos \chi + \frac{\omega^2 r}{V} \cos \varphi \left( \sin \gamma \cos \chi \sin \varphi + \cos \gamma \cos \varphi \right) + 2\omega \sin \chi \cos \varphi$$

(7)

where $m$ is the mass of the spacecraft, and $V$ is the velocity of the spacecraft relative to the atmosphere. $\gamma$ is the flight path angle, $\chi$ is the heading angle, $\lambda$ is the longitude, and $\omega$ is the Earth's angular velocity. Finally, $F_D$ is the drag force, which is computed using Eq. (3).

While the temperature of the object is below the melting temperature of the material, the heat load increases the object temperature. We use a lumped mass schematisation where the object temperature is considered uniform. The temperature variation can be expressed as

$$\frac{dT_w}{dt} = \frac{A_w}{m(t)C_{p,m}} \cdot \left[ \dot{q}_{av} - \varepsilon \sigma T_w^4 \right]$$

(8)

where $T_w$ is the wall temperature, $A_w$ is the wetted area (external area of the object), $m$ is the mass of the object and $C_{p,m}$ is the specific heat at constant pressure. $\dot{q}_{av}$ is the heat flux already adjusted with the shape and attitude dependant averaging factors. $k_{rad}$ is a coefficient that takes into account the heat re-radiation, $\varepsilon$ is the emissivity of the material, and $\sigma$ is the Stefan-Boltzman constant. Once the melting temperature is reached, the object starts melting and losing mass at a rate that is proportional to the net heat flux on the object and the heat of fusion of the material.

$$\frac{dm}{dt} = -\frac{A_w}{h_m} \cdot \left[ \dot{q}_{av} - \varepsilon \sigma T_w^4 \right]$$

(9)

where $h_m$ is the heat of fusion of the material. The material database used is the one available in the NASA Debris Assessment Software (DAS) [17] and integrated with data from [29].

*2.2 Survivability model*

The survivability model analyses the satellite resistance against the impacts of untraceable space debris and meteoroids. Standard vulnerability analysis software relies on ray tracing method in order to predict the damage on spacecraft panels and internal components. Such methods are computationally expensive and require many simulations in order to have a statistically meaningful result, as the impact point of each particle is randomly generated. We here describe a novel methodology that uses a probabilistic approach capable of computing the vulnerability of a spacecraft configuration, avoiding ray tracing methods. The method is based on the concept of vulnerable zone. The vulnerable zone concept is associated to the vulnerability of the internal components of a spacecraft. In fact, it is defined as the area on the external structure of the spacecraft that, if impacted, can lead to an impact also on the component considered. In a simpler fashion, if a space debris impacts on the vulnerable zone of a component, than it also has the possibility to impact such component. The outline of the main characteristics of the software is summarised in the flow diagram of Figure 9.

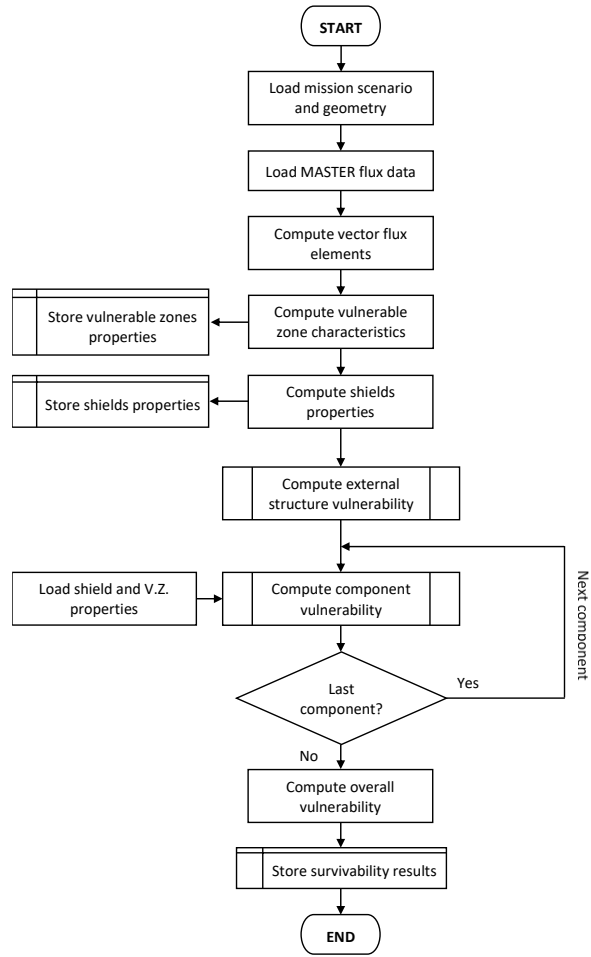

Figure 4: Flow diagram of the main structure of the survivability software.

The procedure involves representing the spacecraft structure with a panelised schematization of its outer structure and using simple shapes for the internal components. To each panel and object, we assign a material selected from a predefined library and geometrical properties such as the type of shielding, the wall thickness, etc.

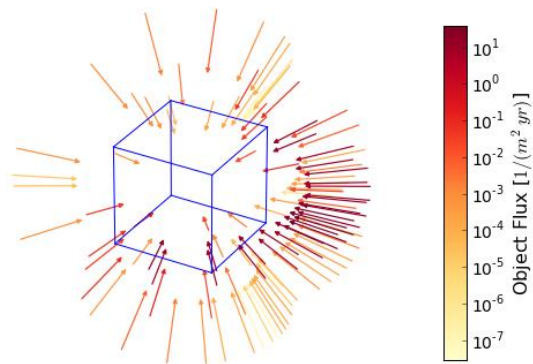

Figure 5: Vector flux elements

Beside the geometrical schematization of the satellite, a representation of the space environment is needed. This is obtained using the European Space Agency (ESA) state of the art software MASTER-2009 [30] that provides a description of the debris environment via flux predictions on user defined target orbit. MASTER-2009 provides a set of 2D and 3D flux distributions as a function of the impact azimuth, impact elevation, impact velocity, and particle diameter. The next step consists in schematizing the debris environment around the satellite using *vector flux elements* [11]. The space around the satellite is subdivided in a set of angular sectors and to each sector is associated a *vector flux element* containing the weighted average of the impact flux, impact direction, impact velocity, and impact diameter [11] (Figure 5). The next step consists in computing the characteristics of the vulnerable zones associated to each internal component. In general, an impact will be oblique (i.e. the angle between the spacecraft face and the impact velocity is not 90 degrees) resulting in two different debris clouds being produced after the impact [31, 32]. One cloud exits almost perpendicularly to the impacted wall, the *normal* debris cloud. The second cloud closely follows the direction of the projectile and is referred to as the *in-line* debris cloud (see Figure 6). We take into account the clouds assuming that the debris belonging to the two clouds are contained inside conic surfaces so that they can be modelled using just the direction of the cone axis and the spread angle of the cone (Figure 6).

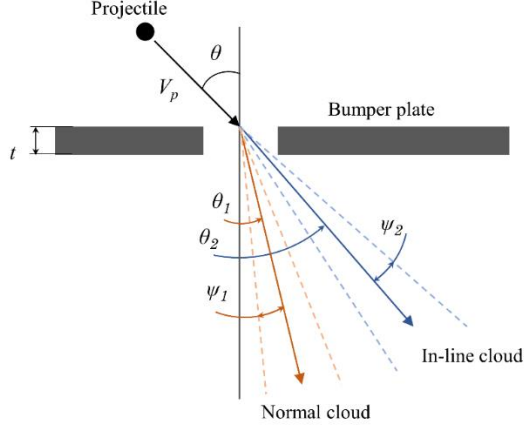

Figure 6: Secondary ejecta clouds characteristics.

The geometry of the cones can be expressed as a function of the impact characteristics (impact velocity, impact angle, particle diameter and wall material) as follows [31, 32]:

$$\frac{\theta_1}{\theta} = 0.471 \cdot \left(\frac{V_p}{C}\right)^{-0.086} \cdot \left(\frac{t}{D_p}\right)^{-0.478} \cdot \cos(\theta)^{0.586}$$

$$\frac{\theta_2}{\theta} = 1.318 \cdot \left(\frac{V_p}{C}\right)^{0.907} \cdot \left(\frac{t}{D_p}\right)^{0.195} \cdot \cos(\theta)^{0.394}$$

$$\tan\psi_1 = 0.471 \cdot \left(\frac{V_p}{C}\right)^{1.096} \cdot \left(\frac{t}{D_p}\right)^{0.345} \cdot \cos(\theta)^{0.738}$$

$$\tan\psi_2 = 1.556 \cdot \left(\frac{V_p}{C}\right)^{-0.049} \cdot \left(\frac{t}{D_p}\right)^{-0.054} \cdot \cos(\theta)^{1.134}$$

(10)

where $t$ is the shield thickness, $D_p$ is the particle diameter, $V_p$ is the particle velocity, and $C$ is the speed of sound in the shield material. $\vartheta$ is the impact angle, $\vartheta_1$ is the normal cloud axis, $\vartheta_2$ is the inline cloud direction, $\psi_1$ is the cone aperture of the normal cloud, and $\psi_2$ is the cone aperture of the inline cloud. These equations have been developed for impacts on Whipple shields and for a range of impact angles ($\theta$) between 30 and 75 degrees. However, it is here assumed that the validity of the equations can be extended to the entire range of impact angles and to other shielding configurations such as honeycomb sandwich panels [9]. The vulnerable zone [9] is defined as the area on the external structure of the spacecraft that, if impacted, can lead to an impact also on the component considered. Any impact of a particle onto this area generate fragments that may hit the component in question, with a probability that depends on the impact parameters, the satellite structure and the stand-off distance of the component from the structure wall. The lateral extent of the vulnerable zone is expressed as

$$2R_{VZ} = 2 \cdot \left[ \tan \alpha_{max} \cdot s + 0.5 \cdot \left( d_{target} + D_{p,max} \right) \right] \tag{11}$$

where (Figure 7) $2R_{VZ}$ is the lateral extension of the vulnerable zone, $s$ is the spacing between the structure wall and the component face (stand-off distance), $d_{target}$ is the lateral size of the considered component, $D_{p,max}$ is the maximum projectile diameter, and $\alpha_{max}$ is the maximum ejection angle. To compute $\alpha_{max}$, a simplification of Eqs. (10) is adopted as it is suggested in [9]. It is assumed that the ejection and spread angles are only a function of the impact angle $\theta$ and that all the other parameters can be absorbed by a constant factor giving:

$$\alpha(\theta) = \theta_2 + \frac{\psi_2}{2} \tag{12}$$

The maximum ejection angle is then $\alpha_{max} = 63.15°$.

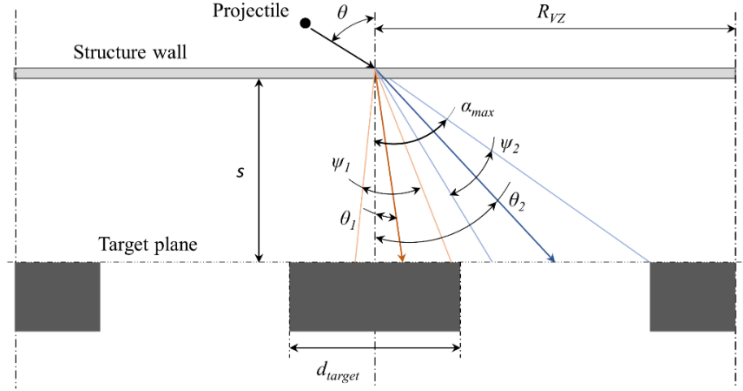

Figure 7: Vulnerable zone extent representation.

The parameter $D_{p,max}$ is a user defined value of the projectile diameter that takes into account the contribution of the particle to the impact probability. Suggested values for $D_{p,max}$ in [9] are 10 mm for vulnerable components and 20 mm for component with higher impact resistance. An example of vulnerable zones as computed by the described procedure is presented in Figure 8.

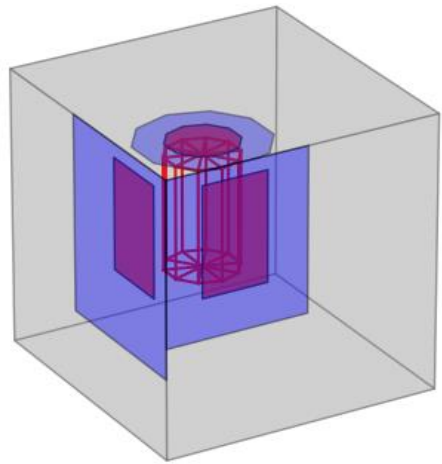

Figure 8: Vulnerable zones of a cylinder projected onto the faces of a cubic structure (closest faces).

After the computation of the vulnerable zones properties, the software automatically detects the characteristics of the shielding, storing all the data that are later necessary for the computation of the probability of penetration. Such data includes the type of panel of the external structure (i.e. single plate or sandwich panel), the thickness of panels, their material, the material of the target object, its thickness, its distance from the respective vulnerable zones, and all the components that can possibly shield the target object.

Once the preliminary operations are terminated, the software analyses the vulnerability of the external structure of the spacecraft. The procedure followed is outlined in the flow diagram of Figure 9.

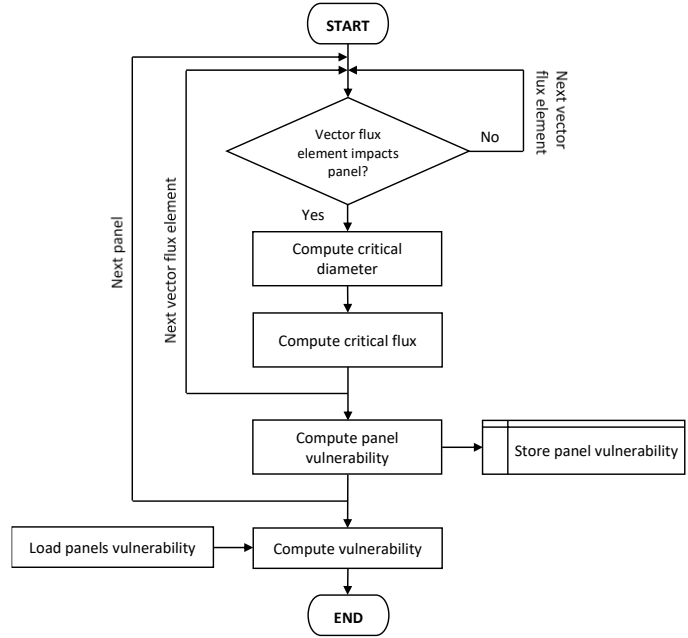

Figure 9: Flow diagram for the vulnerability computation of the external structure.

Once the vector flux elements are computed, the areas of the spacecraft that are susceptible to particle impacts can be determined using a visibility role. Considering the object to be represented by a set $F$ of faces and a vector flux elements with a velocity vector $v_i$, to verify if particles can impact one of the panels in the set $F$ we check the velocity vector $v_i$ with respect to the panel normal $n_j$ using Eq. (13).

$$\mathbf{n_j} \cdot \mathbf{v_i} < 0 \tag{13}$$

All the faces $F$ representing the component are checked against all the vector flux elements following the same visibility role. Then we can compute the probability of such impact. Assuming that debris impact events are probabilistic independent [33], it is possible to use a Poisson statistics to compute the impact probability.

$$P_{imp}^{j,i} = 1 - \exp\left(-\varphi_i \cdot A_\perp^j \cdot t_m\right) \tag{14}$$

where $P_{imp}^{j,i}$ is the probability of an impact on the $j$-th vulnerable zone by the $i$-th vector flux element. $\varphi_i$ is the $i$-th vector flux, $A_\perp^j$ is the projected area of the $j$-th face considered, and $t_m$ is the mission time in years. It is then necessary to compute the penetration probability on the panels. Ballistic limit equations [34] are used to compute the critical diameter using the velocity and direction associated with the vector flux elements together with the geometric and material characteristics of the panels. Once computed the critical diameter for the $i$-th vector flux onto the $j$-th the penetration probability is given by

$$P_p^{j,i} = 1 - \exp\left(-\varphi_{C,i} \cdot A_\perp^j \cdot t_m\right) \tag{15}$$

where $\varphi_{C,i}$ is the particle flux with a diameter greater than the computed critical diameter. With the presented methodology, the computation of the critical flux for each vector flux element replaces the direct sampling of the debris particles. The critical flux can be extracted from the distribution of the cumulative flux vs diameter provided by MASTER-2009 (Figure 10)

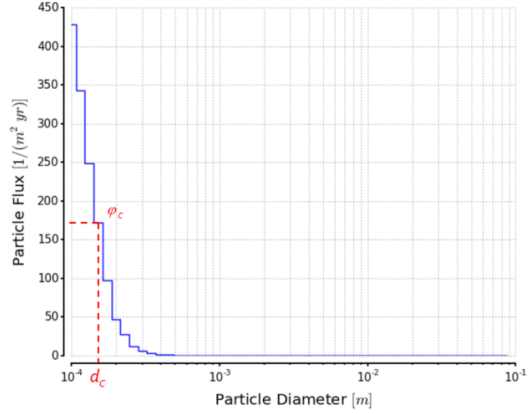

Figure 10: Critical flux computation methodology

As the global distribution of cumulative flux vs diameter is used, the flux extracted is the overall flux for the entire range of azimuth and elevation, it thus cannot be directly used to compute the penetration probability relative to one of the vector flux elements. In fact, to each vector flux is associated a value of the particle flux that is dependent upon the directionality, i.e., impact elevation and impact azimuth, which is a fraction of the total flux. It is here assumed that the distribution of the particles diameter is uniform with respect to the impact direction. With this assumption, the critical flux associated to a vector flux element is considered as a fraction of the overall critical flux. If $\varphi_{TOT}$ is the total debris flux and $\varphi_C$ is the overall critical flux computed, the critical flux relative to the considered vector flux element can be expressed as

$$\varphi_{C,i} = \varphi_i \cdot \frac{\varphi_C}{\varphi_{TOT}} \tag{16}$$

Finally, the penetration probability on the external structure can be computed as follows

$$P_p = 1 - \prod_{j=1}^{N_{panels}} \left( \prod_{i=1}^{N_{fluxes}} \left(1 - P_p^{j,i}\right) \right) \tag{17}$$

with $N_{fluxes}$ total number of vector fluxes elements and $N_{panels}$ total number of panels composing the structure.

After the analysis of the external structure is completed, the internal components are considered. The procedure followed for the internal components is similar to the one for the external structure; however, it directly involves the use of the vulnerable zones. The procedure is schematised in the flow diagram of Figure 11.

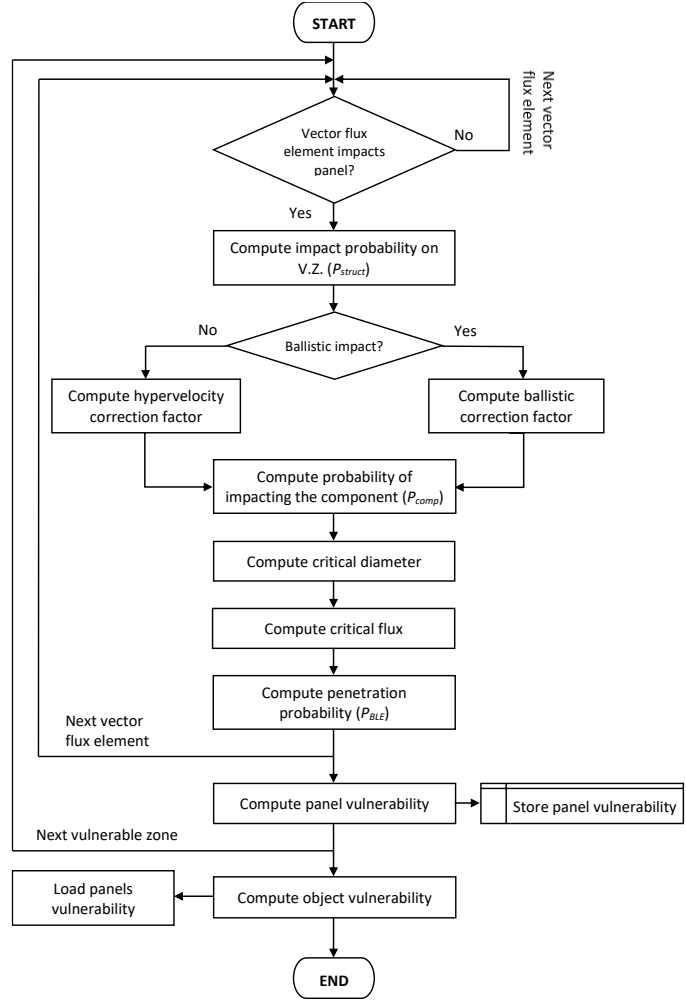

Figure 11: Flow diagram for the vulnerability computation of an internal component.

When a debris particle with sufficient size and velocity impacts onto the outer structure of the spacecraft, a secondary debris cloud is usually generated. Particles belonging to this debris cloud can still impact internal components and damage them. It is thus necessary to evaluate the probability that such secondary debris penetrate the inner components. The vulnerability of an internal component can be evaluated as the product of three different probabilities:

$$P_p = P_{struct} \cdot P_{comp} \cdot P_{BLE} \qquad (18)$$

where $P_{struct}$ is the probability of space debris impacting the spacecraft external structure inside the vulnerable zone assigned to the specific spacecraft component; $P_{comp}$ is the probability that the downrange fragment cloud will hit the component; and $P_{BLE}$ is the probability that the projectile in this cloud perforates the component wall. Consequently, the entire procedure consists in computing the different terms of Eq. (18). The probability of having an impact on the vulnerable zone is similar to the one for the external panels expressed by Eq. (14). The first contribution in the equation $P_{struct}^{j,i}$ can thus be computed as

$$P_{struct}^{j,i} = 1 - \exp\left(-\phi_i \cdot S_{VZ,\perp}^{j} \cdot t_m\right) \qquad (19)$$

where $P_{struct}^{j,i}$ is the probability of an impact on the $j$-th vulnerable zone by the $i$-th vector flux element. $\phi_i$ is the $i$-th vector flux, $S_{VZ,\perp}^{j}$ is the projected area of the $j$-th vulnerable zone corresponding to the component; and $t_m$ is the mission time in years. A particle impacting the vulnerable zone or the resulting debris cloud will not necessarily impact the inner component associated with it. It is thus necessary to take into account the probability that an impact on the vulnerable zone will subsequently cause an impact on the component itself. This is taken into account by the second term on the right end side of Eq. (18). This term ($P_{comp}^{j,i}$) depends on the type of impact: if the impact occurs in the hypersonic regime, secondary debris clouds will form, whereas if the impact is in the ballistic regime the projectile will pass almost intact through the outer structure. It is thus necessary to distinguish between these two situations. In the hypervelocity regime the probability to impact the component is the ratio between the extent of the ejecta in the component plane and the vulnerable zone of the component [9] and can be expressed as:

$$P_{comp}^{j,i} = \frac{d_{ejecta}}{2 R_{VZ}^{j}} \tag{20}$$

where $P_{comp}^{j,i}$ is the probability that the *i*-th vector flux element, which has already impacted the *j*-th vulnerable zone, will hit the component considered. $R_{VZ}^{j}$ is the extent of the *j*-th vulnerable zone in the target plane. $d_{ejecta}$ is the extent of the debris ejecta at the target plane and is computed as follows:

$$d_{ejecta} = 2 \cdot \left[ \tan \alpha_{max}\left(\theta^{j,i}\right) \cdot s_j + 1/2 \cdot d_{target} \right] \tag{21}$$

where $s^j$ is the stand-off distance between the component and the external wall to which the *j*-th vulnerable area is associated, and $\alpha_{max}(\theta^{j,i})$ is the maximum ejection angle associated with the *i*-th vector flux element impacting on the *i*-th vulnerable zone, and can be computed with Eq. (12). In case of an impact in the ballistic regime, only the size of the projectile needs to be considered, as no fragmentation occurs.

$$P_{comp}^{j,i} = \frac{d_p^{j,i} + d_{target}}{2 R_{VZ}^{j}} \tag{22}$$

where $d_p^{j,i}$ is the particle diameter relative to the *i*-th vector flux element impacting on the *j*-th vulnerable zone. This value is associated to each vector flux element and is extracted from the debris flux distributions obtained with MASTER-2009 as the most probable particle diameter for the *i*-th vector flux element. For the scatter regime, a linear interpolation between the ballistic and hypervelocity regime is adopted.

The described procedure still fails to take into account the mutual shielding between components that is the capacity of a component to prevent another one from being hit by the debris particles. In order to take into account such phenomenon, an approach using correction factors has been introduced. Given the different nature of the impacts in the hypervelocity and in the ballistic regime, two correction factors are used. In the case of the hypervelocity regime, the correction factor accounts for the fraction of the extent of the debris ejecta (Eq. (20)) that can be covered by the shielding components. The expression of the correction factor for the *i*-th vector flux elements is as follows:

$$CF_{hyp,i} = 1 - \frac{\sum_{k=1}^{n_s} \left( d_{shield,i} \cdot \frac{s_{target}}{s_{shield,k}} \right)}{d_{ejecta}} \tag{23}$$

where $n_s$ is the number of shielding components between the external face and the target component. $d_{shield,k}$ is the extent of the *k*-th shielding component, $s_{target}$ is the minimum distance of the target component from the external face, $s_{shield,k}$ is the minimum distance between the *k*-th shielding component and the external face, and $d_{ejecta}$ is defined as in Eq. (21). The correction factor takes into account the projection of the extent of each shielding component onto the target plane (Figure 12). In case the extent of $d_{shield}$ exceeds the boundaries of the target plane, $d_{shield}$ is clipped with the target plane, thus only considering the portion of the target projection that is actually inside the spacecraft structure.

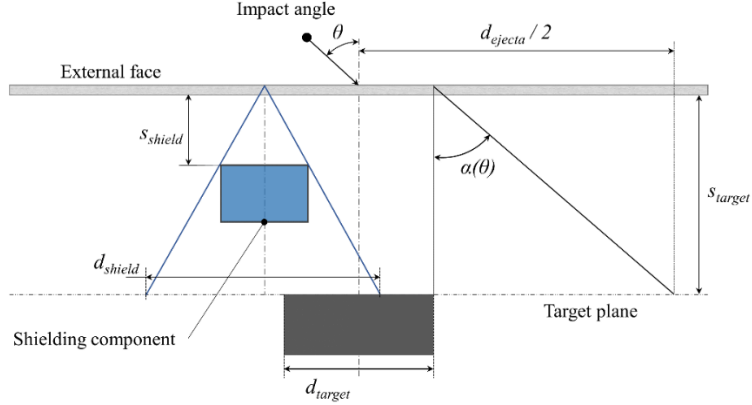
Figure 12: Representation of shielding component contribution to the correction factor.

If the correction factor is 1, no shielding is present, if is 0 the target component is not *visible* by the impactor. It is possible to observe that the correction factor depends on the impact characteristics (i.e. impact angle); therefore, it has to be computed for each vector flux element impacting the considered vulnerable zone, in case the impact is of hypervelocity nature. In the case of the ballistic regime, the same approach cannot be used, as there is no ejecta generation. Looking at Eq. (22), the impact probability in the ballistic regime depends only on the size of the particle and on the extent of the target object. As the dimension of the particle cannot change, we can only act on the extent of the target in order to correct the impact probability. In the case of a ballistic impact we thus use a corrected target extent, which can be referred to as the *visible target extent*. Using an approach similar to the hypervelocity case, the section of the shielding components is projected onto the target plane (Figure 13).

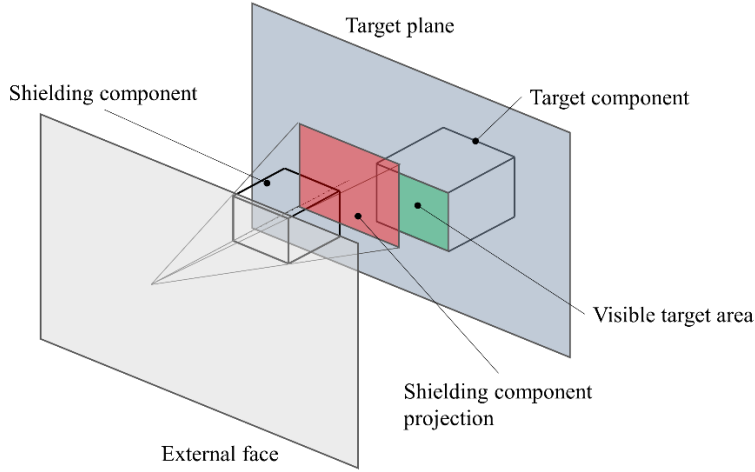
Figure 13: Perspective projection of a shielding component onto the target plane.

At this point, if the projections of the shielding components intersect the target component, they are subtracted from the target components using Boolean operations (difference between sections).

$$A_{v,i} = A_{target} \neg \left( A_{shield,1}, \ldots, A_{shield,k}, \ldots, A_{shield,n_s} \right) \quad (24)$$

Where $A_v$ is the visible area of the target, $A_{target}$ is the target component section, and $A_{shield,k}$ are the areas of the shielding components projected onto the target plane. $\neg$ is the symbol for the Boolean NOT operator, which corresponds to the difference between polygons. Again, whenever a projection of a shielding component has a portion outside the target plane, it is clipped with the target plane itself. As lengths are used in the computation of the impact probability, it is necessary to convert the visible area into a visible length (equivalent). The equivalent visible length ($d_{v,eq}$) is computed as follows:

$$d_{v,eq} = 2 \cdot \sqrt{A_v / \pi} \quad (25)$$

This expression corresponds to the diameter of an equivalent circle with area equal to the visible target area. The ballistic correction factor can then be expressed as

$$CF_{ball} = \frac{d_{v,eq}}{d_{target}} \tag{26}$$

Finally, the hypervelocity and the ballistic can be applied to the impact probabilities of Eqs. (27) and (28) respectively as follows:

$$P_{comp}^{j,i} = \frac{d_{ejecta}}{2 \cdot R_{VZ}^{j}} \cdot CF_{hyp}^{j,i} \tag{27}$$

$$P_{comp}^{j,i} = \frac{d_{p}^{j,i} + CF_{ball} \cdot d_{target}}{2 \cdot R_{VZ}^{j}} \tag{28}$$

The last contribution in Eq. (18) comes from the computation of the penetration probability. The procedure is analogous to the one described for the external structure, and the equation closely resemble Eq. (15).

$$P_{BLE}^{j,i} = 1 - \exp\left(-\phi_{C,i} \cdot S_{VZ,\perp}^{j} \cdot t_m\right) \tag{29}$$

where $P_{BLE}^{j,i}$ is the penetration probability for the *j*-th vector flux element on the component associated with the *i*-the vulnerable zone, and $\phi_{C,i}$ is the critical flux that is the flux associated to the value of the critical diameter computed with ballistic limit equation. The adopted ballistic limit equation is the Schafer-Ryan-Lambert (SRL) [9, 34]. The SRL BLE is a triple-wall ballistic limit equation and can be used for both triple-wall and double wall-configuration. The software uses this equation in a way that assumes the last wall of the shielding configuration is always the face of the inner component considered, whereas the other walls represent the outer structure. The overall penetration probability for a component can then be computed combining al the previous contributions as follows

$$P_p = 1 - \prod_{j=1}^{N_{panels}} \left( \prod_{i=1}^{N_{fluxes}} \left(1 - P_{struct}^{j,i} \cdot P_{comp}^{j,i} \cdot P_{BLE}^{j,i}\right) \right) \tag{30}$$

## 3 Sun-synchronous missions

As mentioned, for the presented analysis we are considering the optimisation of tank configurations of Earth observation and remote sensing missions [35, 36]. These kinds of missions frequently exploit sun-synchronous orbits and that is the reason why the current work focuses on these orbits. A sun-synchronous orbit is a Low Earth Orbit (LEO) that combines altitude and inclination in order for the satellite to pass over any given point of the Earth's surface at the same local solar time, granting the satellite a view of the Earth's surface at nearly the same illumination angle and sunlight input. To estimate the size of the tank assembly it is necessary to compute the amount of propellant needed for the mission through a delta-V budget. As sun-synchronous orbits are influenced by atmospheric drag and by the non-uniformity of the Earth's gravitational field, they require regular orbit correction manoeuvres. They also need, as for most spacecraft, additional manoeuvres to correct orbit injection errors and to perform disposal manoeuvres. The computation of the different contributions to the delta-V budget is described in the following paragraph.

*3.1 Delta-V budget*

To estimate the tankage volume it is necessary to compute the amount of propellant needed by the spacecraft as a function of the mission characteristics. The three main elements that contribute to the ΔV budget for the required mission lifetime are the orbit maintenance, the launch injection errors, and the disposal manoeuvres.

Orbit maintenance manoeuvres are used to maintain the sun-synchronism of the orbit and to control the ground track with a given accuracy. To do so, the orbital height and inclination need to be maintained within admissible ranges. In LEO, atmospheric drag results in orbital decay, causing the semi-major axis and the orbit period to decrease. The reduction in the semi-major axis $\delta a$ and in the orbital period $\delta \tau$ for one orbit can be computed as

$$\delta a = -2\pi \rho_{atm} \frac{SC_D}{m_s} a_0^2 \tag{31}$$

$$\delta\tau = \frac{3\pi}{V_s}\delta a \qquad (32)$$

where $\rho_{atm}$ is the atmospheric density, $S$ is the average cross section of the satellite, $C_D$ is the drag coefficient, $m_s$ is the mass of the satellite, $a_0$ is the nominal orbit semi-major axis, and $V_s$ is the orbital velocity of the spacecraft. The changes in the orbital height and period lead to changes in the ground track. Such variations can be controlled by imposing a tolerance on the nominal ground track. When the spacecraft's ground track reaches the prescribed tolerance, a correction manoeuvre needs to be executed. To do so, the time difference from the nominal time at the equator passage $\Delta t_0$ needs to be computed:

$$\Delta t_0 = \frac{\Delta\lambda}{\omega_e} \qquad (33)$$

where $\omega_e$ is the angular speed of the Earth and $\Delta\lambda$ is the longitude displacement at equator passage and can be expressed as:

$$\Delta\lambda = \frac{2E_0}{r_e} \qquad (34)$$

$r_e$ is the radius of the Earth, and $E_0$ is the imposed tolerance on the displacement from the nominal orbit ground track at the equator (equal to 0.7 km for this study). Using Eqs. (33) and (34) It is possible to compute the number of orbits after which the equator crossing displacement reaches the prescribed limit as follows:

$$k = \sqrt{\frac{2\Delta t_0}{\delta\tau}} \qquad (35)$$

To control the ground track, the manoeuvre has to be executed every $2k$ orbits, leading to a variation in the orbit semi-major axis ($\Delta a_{decay}$) and orbital period ($\Delta t_{decay}$) of:

$$\begin{aligned}\Delta a_{decay} &= 2k|\delta a| \\ \Delta t_{decay} &= 2k|\delta\tau|\end{aligned} \qquad (36)$$

$\Delta t_{decay}$ is also the time between the necessary orbit correction manoeuvres. The correction manoeuvre can be computed with a Hohmann transfer:

$$\Delta V_{decay,i} = \sqrt{\frac{\mu_e}{r_1}}\left(\sqrt{\frac{2r_2}{r_1+r_2}}-1\right) + \sqrt{\frac{\mu_e}{r_2}}\left(1-\sqrt{\frac{2r_1}{r_1+r_2}}\right) \qquad (37)$$

where $\mu_e$ is the gravitational parameter of the Earth, $r_1 = a_0 - \Delta a_{decay}$ is the radius of the initial circular orbit, and $r_2 = a_0$ is the radius of the final orbit after the manoeuvre. The total $\Delta V_{decay}$ due to the orbital height correction manoeuvres for the entire mission lifetime is the sum of the contribution of Eq. (37) every $\Delta t_{decay}$ so that:

$$\Delta V_{decay} = \left\lfloor\frac{t_m}{\Delta t_{decay}}\right\rfloor \Delta V_{decay,i} \qquad (38)$$

where $\lfloor t_m / \Delta t_{decay} \rfloor$ represents the number of manoeuvres to be executed during the mission lifetime $t_m$.

In addition, the orbit inclination needs to be controlled during the lifetime of a sun-synchronous spacecraft. The variation of the orbital inclination in fact causes the drifting of the line of the nodes and affects ground track repetition. The total $\Delta V_{inc}$ needed to compensate for the inclination variation can be computed as:

$$\Delta V_{inc} = 2\sin\left(\frac{\Delta i_{sec}}{2}\right)\cdot t_m \qquad (39)$$

where $\Delta i_{sec}$ is the secular variation of the inclination in one year that can be assumed equal to 0.05 deg/year, and $t_m$ is the mission time in years.

To compute the $\Delta V_{inj}$ needed to compensate for injection errors, we assume that the maximum errors in the orbital parameters after launch are:

$$\Delta a_{inj} = \pm 35 \text{ km}$$
$$\Delta i_{inj} = \pm 0.2 \text{ deg} \qquad (40)$$

The $\Delta V_{inj}$ due to the injection errors can then be computed using a Hohmann transfer with plane change where the initial and final orbits have a radius of $r_1 = a_0 - \Delta a_{inj}$ and $r_2 = a_0$ respectively, and the inclination change is equal to $\Delta i_{inj}$. Finally, the $\Delta V_{disp}$ to ensure the end-of-life disposal of the satellite can be computed as follows. It is possible to consider as a disposal manoeuvre a Hohmann transfer from the nominal orbit to a 600 km orbit, assuming that the 600 km altitude will allow a spacecraft to decay naturally within 25 years,

The sum of the previously computed delta-V values is the total $\Delta V_{tot}$ budget of a sun-synchronous mission, which depends on the nominal orbit of the spacecraft, the mission duration, and the characteristics of the spacecraft (mass, cross-section, drag coefficient).

$$\Delta V_{tot} = \Delta V_{decay} + \Delta V_{inc} + \Delta V_{inj} + \Delta V_{disp} \qquad (41)$$

*3.2 Tankage volume*

For the purpose of this work, it is assumed that a monopropellant hydrazine propulsion system is adequate for all the orbit correction manoeuvres previously described. The specific impulse of hydrazine is 200 s. The propellant mass needed by the spacecraft during its entire lifetime can be computed using the Tsiolkowsky equation [35].

$$m_f = m_{s,in}\left(1 - e^{-\frac{\Delta V_{tot}}{g_0 Isp}}\right) \qquad (42)$$

where $m_f$ is the propellant mass needed to perform the total velocity change $\Delta V_{tot}$, $m_{s,in}$ is the initial spacecraft mass, $g_0$ is the gravitational acceleration at sea level (equal to 9.81 m/s$^2$), and $Isp$ is the specific impulse of the fuel used.

Once the propellant mass is calculated, the tankage volume can be estimated as

$$v_p = K1 \cdot \frac{m_f}{\rho_f} \qquad (43)$$

where $\rho_f$ is the density of hydrazine (equal to 1.02 g/cm$^3$), and $K1$ is a factor that takes into account the additional volume needed for the pressurant gas. For the entire article, $K1$ is assumed to have a value of 1.4 (average value from [37]). As an example, let us consider the MetOp mission [38]. MetOp is a sun-synchronous satellite with a mass of 4085 kg, and an average cross section $S = 18$ m$^2$. The operational orbit of the mission is 817 km in altitude with an inclination of 98.7 degrees. The mission design life is 5 years. Computing the mass of propellant with Eq. (42) returns a value of 360 kg of propellant, which is 12.5% more propellant than the actual mission of 320 kg. This is a reasonably close value considered considering the approximate delta-V budget procedure. Moreover, the value used for the specific impulse in the article (200 s) is lower than the actual characteristics of the MetOp mission thrusters, which ranges between 220 s and 230 s. Using these two values, the resulting propellant mass would range between 332 kg and 321 kg, a much closer value to the original mission. As another example, Cryo-Sat2 [39] is a 3 years mission with a satellite mass of 720 kg, an average cross section of 8.8 m$^2$, and an orbital altitude of 717 km. The resulting propellant mass is 43 kg that is in good agreement with the value of 38 kg of the actual mission.

## 4 Multi-objective optimisation

The demisability and survivability models have been implemented into a multi-objective optimisation framework. The purpose of this framework is to find preliminary, optimised spacecraft configurations, which take into account both the survivability and the demisability requirements. In this way, a more integrated design can be achieved from the early stages of the mission design. The requirements arising from the demisability and the survivability are in general competing; consequently, optimised solutions represent trade-offs between the two requirements. In its most general form, a multi-objective optimisation problem can be formulated as:

$$\begin{aligned}
\text{Min/Max } & f_m(\mathbf{x}), & m &= 1, 2, ..., M; \\
\text{Subject to } & g_l(\mathbf{x}) \leq 0, & l &= 1, 2, ..., L \\
& h_b(\mathbf{x}) = 0, & b &= 1, 2, ... B; \\
& x_i^{(L)} \leq x_i \leq x_i^{(U)}, & i &= 1, 2, ..., n.
\end{aligned} \quad (44)$$

where $\mathbf{x}$ is a solution vector, $f_m$ is the set of the $m$ objective functions used, $g$ and $h$ are the constraints and $x_i^{(l)}$ and $x_i^{(U)}$ are the lower and upper limits of the search space. In multi-objective optimisation, no single optimal solution exists that can minimise or maximise all the objective functions at the same time. Therefore, the concept of Pareto optimality needs to be introduced. A Pareto optimal solution is a solution that cannot be improved in any of the objective functions without producing a degradation in at least one of the other objectives [40, 41]. There exists a large variety of optimisation strategies; however, for the purpose of this work and for the characteristics of the problem in question, genetic algorithms have been selected. The Python framework Distributed Evolutionary Algorithms in Python (DEAP) [42] was selected for the implementation of the presented multi-objective optimisation problem. Specifically, the selection strategy used is the Non-dominated Sorting Genetic Algorithm II (NSGAII) [41]. For the crossover mechanism, the Simulated Binary Bounded [43] operator was selected whereas for the mutation mechanism the Polynomial Bounded [44] operator was the choice. Throughout this article, the input parameters to the genetic algorithm that define the characteristics of the evolution were fixed: the size of the population was set to 80 individuals, and the number of generations was set to 60. The crossover and mutation probability were 0.9 and 0.05 respectively.

*4.1 Optimisation variables and constraints*

The variables (vector $\mathbf{x}$ in Eq. (44)) of the optimisation process relate to the internal components, specifically the propellant tanks. The variables to be optimised were the tank material, the thickness, the shape, and the number of tanks. The total tankage volume, which in turn influences the internal radius of the tanks, is determined using Eq. (43) and depends on the mission scenario.

For the material, the possible options are limited to three different materials typically used in spacecraft tank manufacturing. The possible materials are aluminium alloy Al-6061-T6, titanium alloy Ti-6Al-4V, and stainless steel A316. The characteristics of these materials are summarised in Table 4.

Table 4: Properties of the materials used in the optimisation [45, 46].

|  | Al-6061-T6 | A316 | Ti-6Al-4V |
|---|---|---|---|
| $\rho_{mat}$ (kg/m$^3$) | 2713 | 8026.85 | 4437 |
| $T_m$ (K) | 867 | 1644 | 1943 |
| $C_m$ (J/kg-K) | 896 | 460.6 | 805.2 |
| $h_f$ (J/kg) | 386116 | 286098 | 393559 |
| $\varepsilon$ | 0.141 | 0.35 | 0.3 |
| $\sigma_y$ (MPa) | 276 | 415 | 880 |
| $\sigma_u$ (MPa) | 310 | 600 | 950 |
| $C$ (m/s) | 5100 | 5790 | 4987 |

The shape of the tanks can take two different geometries: a spherical tank or a right cylindrical tank, (represented in the optimisation using a binary value). These geometries were chosen because they are typical of actual tank designs. The number of tanks in which the propellant can be divided was varied from one to six units. It was assumed that six tanks would be a reasonable upper limit for the possible number of tanks to adopt.

Lastly, the thickness of the tanks can be varied in the range 0.5 mm to 5 mm. This was considered a reasonable range for actual spacecraft tanks. Values smaller than 0.5 mm are considered too small, and more suitable for tank liners. Values larger than 5 mm were excluded because very thick metallic tanks would be too heavy. A summary of the variables of the optimisation with their respective values and range is provided in Table 5

Table 5: Summary of optimisation variables.

| Variable | Range/Values | Variable type |
|---|---|---|
| Tank material | Al-6061-T6, Ti-6Al-4V, A316 | Integer |
| Tank number | 1 to 6 | Integer |

| Tank thickness | 0.0005 to 0.005 m | Real |
| Tank shape | Sphere, Cylinder | Integer |

*4.2 Definition of fitness functions*

The developed multi-objective optimisation framework uses the demisability and survivability models (see Section 2) to compute the fitness functions. These fitness functions allow the evaluation of a certain spacecraft configuration against the demisability and survivability. In addition, they allow the comparison between different solutions, giving the optimiser the possibility to rank the different solutions according to their demisability and survivability.

To evaluate the level of demisability of a certain configuration, the Liquid Mass Fraction (LMF) is introduced. The LMF index represents the proportion of the total re-entering mass that melts during the atmospheric re-entry. In mathematical terms, the LMF index can be expressed as:

$$LMF = 1 - \frac{\sum_{j=1}^{N} m_{fin,j}}{\sum_{j=1}^{N} m_{in,j}} \qquad (45)$$

where $m_{fin,j}$ and $m_{in,j}$ are the final and initial mass of the $j$-th component respectively, and $N$ is the total number of components.

To evaluate the level of survivability, the probability of no-penetration (PNP) was selected as the survivability fitness function. The probability of no-penetration represents the chance that a specific spacecraft configuration is not penetrated by space debris during its mission lifetime. In this case, the penetration of a particle is assumed to produce enough damage to the components to seriously damage them so that the PNP can be considered a sufficient parameter to evaluate the survivability of a satellite configuration. The overall probability of no-penetration of a spacecraft configuration is given by:

$$PNP = 1 - \sum_{j=1}^{N} P_{p,j} \qquad (46)$$

where $P_{p,j}$ is the penetration probability of the $j$-th component. Strictly speaking, the $i$ index could produce values below 0. However, it is possible to observe from several impact analysis on satellite configurations [10, 47-49] that the penetration probability on internal components is usually low. For a single internal component, it is usually under 1%. Consequently, in practical cases, a value below zero is not to be expected from the *PNP* index.

*4.3 Optimisation setup*

As the demisability and the survivability are complex and require many different inputs, it is necessary to take into account all these different parameters. It is not only necessary to define the variables of the optimisation (see Section 4.1), but also all the other parameters needed by the two models to carry out their computations. One of the main aspects to define is the mission scenario (see Section 4.3.1) for both the demisability and the survivability. In the first case, this means taking a decision about the initial conditions of the atmospheric re-entry. In the second case, the operational orbit of the satellite and the mission duration need to be defined.

In the present work, the objective of the optimisation is to optimise tank configurations. Two aspects of the tank configuration that are not directly taken into account by the optimisation variables is the size and position of the tanks (see Section 4.3.3). It was decided to relate the size of the tanks, i.e. the radius, to the total tankage volume (see Section 3.2) and to the number of tanks (see Eqs. (49) and (50)) in order to have a more realistic mission scenario. Delta-V budgets are in fact one of the main constraints on the mission design process and the amount of propellant, which is related to the size of the tanks, needs to be sufficient for the mission requirements. For what concerns the tank configurations, it was decided that, for the current stage of development of the project, the introduction of the optimisation of the positions of the tanks inside the spacecraft would have been too complex. For this reason, a predefined set of positions for the tanks was adopted (see Section 4.3.3).

Finally, both the demisability and the survivability analysis cannot be carried out without knowing the characteristics of the main spacecraft structure, i.e. the overall size and mass of the satellite, the material, the thickness, and the type of shielding (see Section 4.3.2).

*4.3.1 Mission scenarios*

For the demisability simulation, the initial re-entry conditions are represented by the altitude, the flight path angle, the velocity, the longitude, the latitude, and the heading angle. Standard values for these parameters [50, 51] were selected and are presented in Table 6.

Table 6: Initial conditions for the re-entry simulations.

| Parameter | Symbol | Value |
|---|---|---|
| Altitude | $h_{in}$ | 120 km |
| Flight path angle | $\gamma_{in}$ | 0 deg |
| Velocity | $v_{in}$ | 7.3 km/s |
| Longitude | $\lambda_{in}$ | 0 deg |
| Latitude | $\varphi_{in}$ | 0 deg |
| Heading | $\chi_{in}$ | -8 deg |

For the survivability, the mission scenario is defined by the operational orbit of the satellite. As previously introduced (see Section 0), we are considering sun-synchronous missions. For this reason, the orbits selected need to satisfy the sun-synchronicity requirement. Specifically, orbits with an inclination of 98 degrees and an altitude of 800 km were chosen. In addition, four different mission durations were selected: 3, 5, 7, and 10 years.

*4.3.2 Spacecraft configuration*

The variables of the optimisation are related to internal components; however, the external configuration of the satellite still needs to be defined in order to perform the demisability and the survivability analysis. The first decision concerns the shape and the dimension of the outer structure of the satellite. It was decided to adopt a cubic shaped spacecraft in order to keep the analysis as general as possible. The dimensions of the cubic structure (i.e. its side length) can be computed tacking into account the mass of the satellite ($m_s$) and assuming an average density for the satellite ($\rho_s$) as follows [35].

$$L = \sqrt[3]{m_s / \rho_s} \tag{47}$$

It was assumed that the density of the spacecraft is 100 kg/m³, which is an average value that can be used in preliminary design computations [35].

Four classes of satellites were considered in the present analysis. The classes were defined according to the mass of the spacecraft: 500 kg, 1000 kg, 2000 kg, and 4000 kg options were considered. The classes and the corresponding spacecraft size, computed with Eq, (47) are summarised in Table 7.

Table 7: Mission classes analysed with respective size of the satellite.

| Class | Side length |
|---|---|
| 500 kg | 1.7 m |
| 1000 kg | 2.15 m |
| 2000 kg | 2.7 m |
| 4000 kg | 3.4 m |

In addition to the size and mass of the spacecraft, the thickness and material of the external wall also need to be defined. For the purpose of this work, and in order to maintain the same conditions for all the simulations, it was decided to use a single wall configuration with a 3 mm wall thickness made of Aluminium alloy 6061-T6.

*4.3.3 Tank configurations*

To describe the possible tank configurations, a series of assumptions were made in order to limit the complexity of the problem so that it could be analysed within the current capabilities of the demisability and survivability models developed (Section 2). The first assumption was to limit the maximum number of tanks to six units (i.e. the total propellant mass cannot be divided into more than six tanks). The second assumption concerns the disposition of the tanks inside the spacecraft. Because of the limitations on the position of the centre of mass of the satellite, it was decided to equally space the tanks around the centre of mass. The centre of each tank is placed at the vertices of a regular polygon and the barycentre of the polygon coincides with the centre of the main spacecraft structure. For example, three tanks would be positioned as an equilateral triangle, four tanks as a square, and so on. As the tanks obviously cannot intersect each other, their mutual distance has to be bigger than twice their radius. With this consideration, the side length of the polygons can be computed as:

$$l = 2r_t \cdot K2 \tag{48}$$

where $r_t$ is the tank radius, and $K2$ is a multiplicative factor to take into account the spacing between two tanks. For the analysis presented in this paper, $K2$ has a value of 1.2.

As the total tankage volume is fixed by the mission characteristics and computed through Eq. (43), the external radius of the tank can be related to the number of tanks in the configuration as follows. For spherical tanks, we have:

$$r_t = \sqrt[3]{\frac{3}{4\pi} \cdot \frac{v_p}{n_t}} + t_t \tag{49}$$

Whereas for right cylindrical tanks

$$r_t = \sqrt[3]{\frac{1}{2\pi} \cdot \frac{v_p}{n_t}} + t_t \tag{50}$$

where $r_t$ is the outer radius of the tank, $n_t$ is the number of tanks in the configuration, and $s_t$ is the thickness of the tank. An example of a configuration with four tanks is presented in Figure 14.

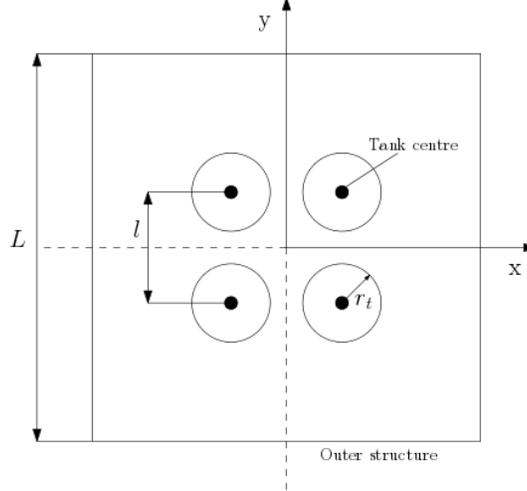

Figure 14: Example of a tank configuration with four tanks equally spaced with respect to the centre of mass of the spacecraft.

## 5 Results and discussion

This section presents the results obtained through the multi-objective optimisation framework previously described (see Section 4). The results were used to analyse the influence of the spacecraft design and of the mission characteristics on the demisability and the survivability when these factors are considered in combination. The influence of the mission characteristics in term of the mass of the spacecraft (see Table 7) and the mission lifetime were taken into account. The effect of the tank configuration with respect to the tank material, size, thickness, shape, and number of vessels was also considered.

The analysis considered cylindrical and spherical tanks with varying thickness, material, and number of vessels. The possible tank configurations ranges from one to six vessels and are positioned in a simplified fashion, as shown in Figure 15. The circles represent the positions of the centre of each tank. The size and positioning of the tanks follows the procedure of Section 4.3.3. In each configuration, the tanks have the same shape, i.e. they are all spheres or cylinders.

Page 24 of 36

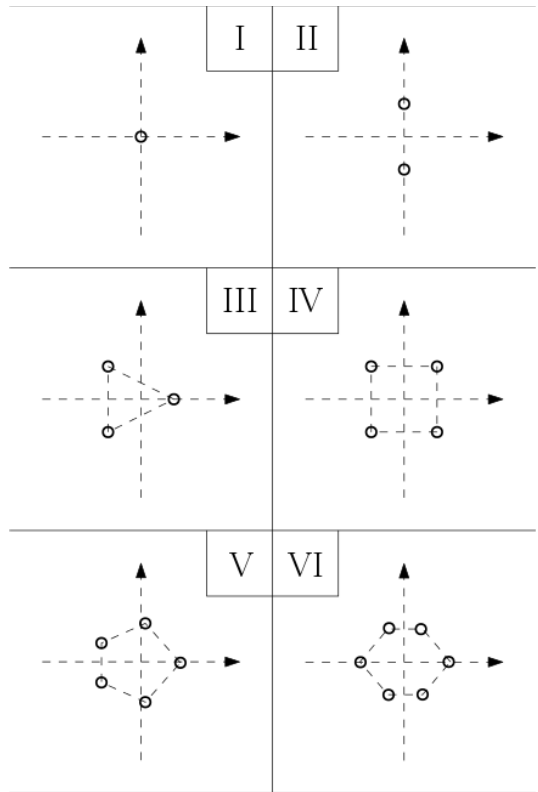

Figure 15: Possible configuration for the positioning of the spacecraft tanks.

Different optimisation simulations were performed varying the range of the number of tanks, i.e. not just performing the simulations with the maximum possible number of tanks.

An example of the Pareto front obtained with the optimisation is shown in Figure 16. The presented Pareto front was obtained for a 2000 kg spacecraft in an 800 km altitude orbit, with a maximum number of tanks allowed equal to three (configurations I, II, and III of Figure 15). The x-axis represents the survivability index, i.e. the Probability of No-Penetration, and the y-axis represents the demisability index, i.e. the Liquid Mass Fraction. Both indices are expressed in terms of percentage. It is possible to observe the general trend of the Pareto front, with the aluminium solutions in the upper part (red solutions), representing the solutions with higher demisability but also a higher vulnerability to debris impacts. The stainless steel solutions (blue solutions) are instead on the right part of the graph, corresponding to solutions with higher survivability but lower demisability. No titanium solutions are present. This is a common result for all the simulations performed, due to the extremely low demisability of titanium and its impact resistance comparable to stainless steel.

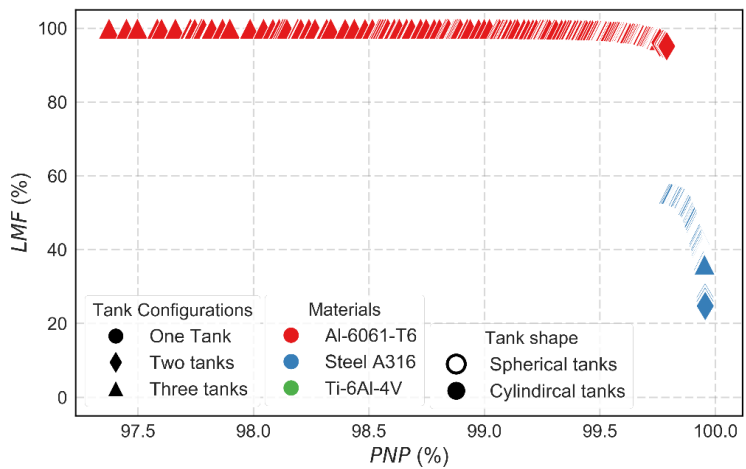

Figure 16: Pareto front for a 2000 kg spacecraft and 10-year mission with a maximum allowed number of tanks equal to three.

A gap is also observable between the aluminium and the stainless steel solutions. This is caused by the considerable difference between the demisability of the two materials. In fact, two solutions with a slightly different survivability coming from the different combination of material and thickness can have a remarkable difference in the demisability because of the high influence of material properties on the demisability. All the solutions obtained by the optimiser represent configurations with cylindrical tanks. In general, no solutions with spherical tanks were obtained for the conditions used in this study. Another interesting consideration is the influence of the number of tanks in the configuration. As it is possible to observe, in this case, all three configurations are viable solutions after the optimisation. Configurations with lower amount of tanks have higher survivability and the lower demisability. This is because the smaller the number of tanks in which to split the propellant, the bigger the tanks are and thus the less demisable they are. However, they also have a lower external surface, which in turn means a higher survivability. On the other hand, configurations with a higher number of tanks have smaller and more demisable tanks but a higher external surface. They are also positioned closer to the external walls making them more vulnerable to the debris impacts.

Figure 17 and Figure 18 represent two further examples of Pareto fronts for the previously introduced mission scenario. In these cases, the maximum allowed number of tanks was respectively four and six. As it is possible to observe in Figure 17 the only solutions resulting from the optimisation are those corresponding to Configuration IV, with four tanks. Figure 18 has more variety of solutions with configurations comprising four, five, and six tanks. However, there is a clear predominance of solutions with six tanks. In general, in all the optimisations performed, for the different classes and mission scenarios, this trend was repeated.

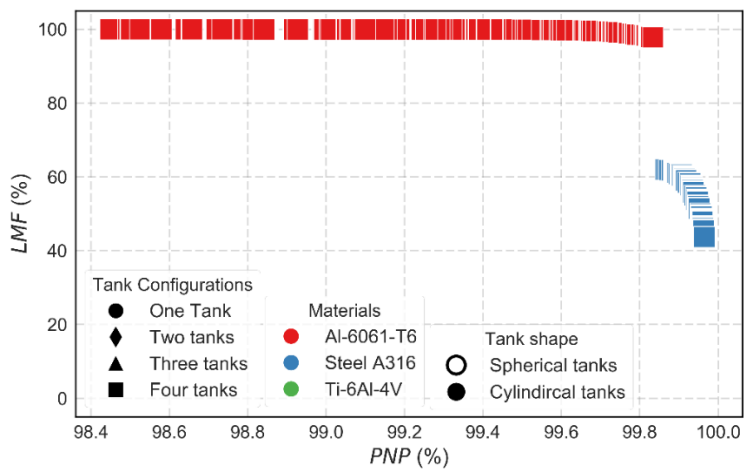

Figure 17: Pareto front for a 2000 kg spacecraft and 10-year mission with a maximum allowed number of tanks equal to four.

The majority of the solutions were represented by the maximum allowed number of tanks. Despite the fact that a higher number of tanks is also more exposed to debris impact, it is possible for the optimiser to find solutions with a higher thickness to compensate for this while still maintaining a higher demisability than a solution with a lower number of tanks. When this is no longer the case within the specified ranges of the optimisation, solutions with a lower number of tanks become better. This behaviour is strictly correlated with how the demisability index is defined (see Eq. (45) where the proportion of mass demised is considered. This is a reasonable choice for a demisability index. In fact, while it is true that even a partially melted object reaches the ground and contributes to the casualty risk, the uncertainty in re-entry simulations [52, 53] make the LMF index a good indication of how likely the considered object will demise. It is in fact important to remember that the presented methodology involves a preliminary assessment of many configurations. A further, more refined, analysis of the most promising configurations can then be carried out to verify their quality. Another observable trend within Figure 16, Figure 17, and Figure 18 is the reduction of the demisability gap between the aluminium alloy and the stainless steel solutions. As the number of tanks increases, the maximum demisability of the stainless steel solutions also increases, closing the gap with the aluminium alloy solutions. This indicates that to have demisable solutions for tanks made of stainless steel, it is necessary to increase the number of vessels used, as only smaller tanks will be demisable.

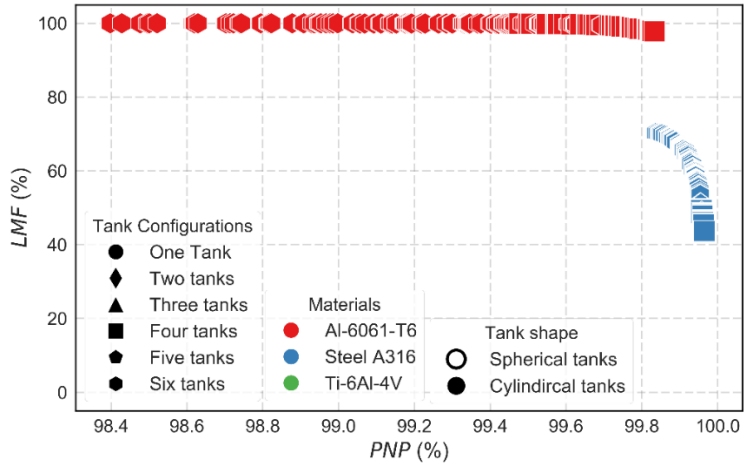

Figure 18: Pareto front for a 2000 kg spacecraft and 10-year mission with a maximum allowed number of tanks equal to six.

As introduced in the previous paragraphs, the mission characteristics have also been taken into account. In particular, the size of the spacecraft in terms of mass and external dimension (see Table 7), and the mission lifetime. These aspects influenced the optimisation results, changing the shape and the features of the Pareto fronts. For example, the minimum value of the PNP index corresponds to the configuration with the best demisability but also the worst survivability. Such information gives an idea of how much an improved demisability can compromise the survivability of a spacecraft, and as a function of the mission scenario. In Figure 19 it is possible to observe the variation of the minimum survivability for each optimisation simulation as a function of the mission class and lifetime, and for four different tank configurations.

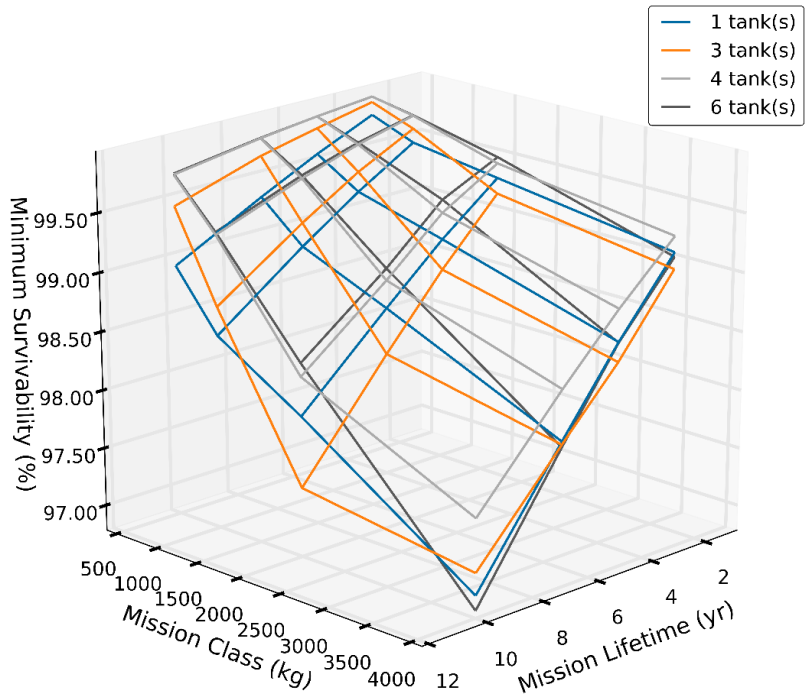

Figure 19: Minimum survivability of the solutions in the Pareto front as a function of the mission class and lifetime.

As expected, the higher the mission lifetime, the lower the value of the minimum survivability. The same happens as we increase the mission class. In fact, bigger spacecraft have a bigger probability of being hit by space debris. The figure shows that for smaller spacecraft with limited lifetimes an enhanced demisability may not compromise their survivability, giving the design team the possibility to push to more extreme solutions favouring the demise. On the other hand, more massive and longer mission may need a more refined study because very demisable solutions may compromise the reliability of the mission with respect to debris impacts. The difference is in fact in the order of 2.5%, which is not negligible. In fact, the penetration probability of complete spacecraft configurations is usually in the order of few percentage points (1-5%) [54]. Therefore, having the survivability of just one component hampered by 1% or 2% would have a big relative impact on the overall survivability of a spacecraft. It is true that this is a simplified simulation and many things can be changed in a spacecraft configuration to make it less vulnerable, for instance the outer shielding. However, this study gives a first order evaluation of what the effects can be when the survivability and the demisability are considered at the same time.

Another feature of the Pareto front that is heavily influenced by the simulation class and mission duration is the demisability gap that is present between the aluminium solution with the smallest demisability and the steel solution with the highest demisability. This gap provides an indication of how difficult it is for steel solutions to reach a comparable demisability level with respect to aluminium solutions. Figure 20 shows a representation of the trend relating to this demisability gap. As it is possible to observe, the higher the number of tanks in the configuration, the smaller the gap for every type of mission scenario.

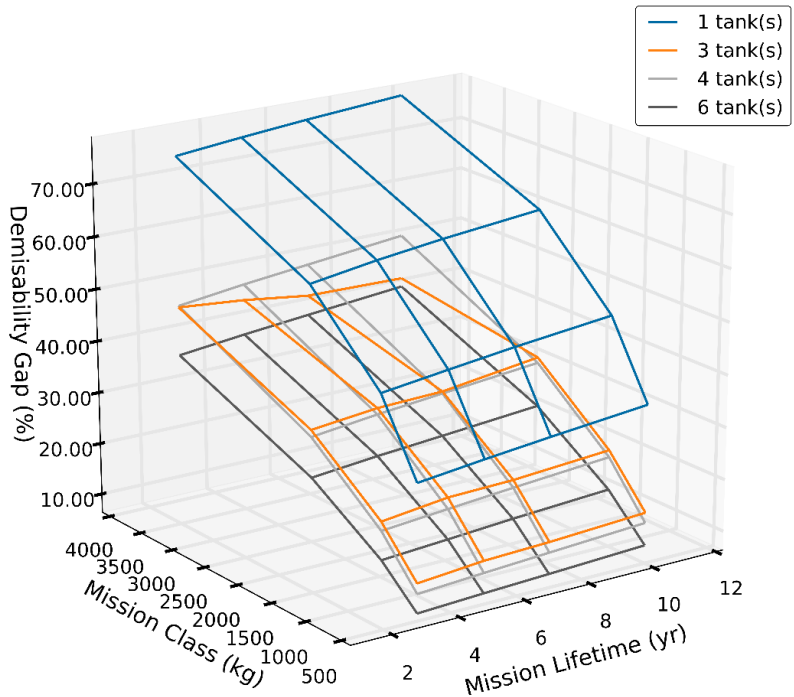

Figure 20: Demisability gap between the solutions in the Pareto front as a function of the mission class and lifetime.

There is a considerable gap between the solution with one tank and the other solutions with multiple tanks. It is also possible to observe that the configurations with three and four tanks are very close and produce similar results. It is also evident that the mission size is much more influential than the mission lifetime for this specific characteristic. Bigger missions have larger tanks, which are of course more difficult to demise. This has a bigger impact on solutions with less demisable materials such as the stainless steel.

### 5.1 Optimisation indices discussion

The optimisation framework presented in the previous paragraphs was aimed at demonstrating the interdependence of the demisability and the survivability requirements when it comes to the design of spacecraft components and configurations. To do so, a simplified spacecraft design with a single type of internal component (i.e. tanks) was selected. In order to relate the results obtained to actual mission scenarios, the propellant mass and, as a consequence, the dimensions of the tanks was related to the delta-V budget of sun-synchronous missions (see Section 0). The optimiser is free to look for solutions inside the search space defined in Table 5. No other constraints were added to the optimisation problem. For example, there was no limit on the overall mass of the configuration or complexity of the propulsion regulation system, and no check if the tanks were sufficiently strong to withstand the common storage pressures, etc. It is thus interesting to compare the obtained solutions against some of these characteristics. In Figure 21 a Pareto front is presented for a 2000 kg class mission with seven years of mission lifetime and a maximum allowed numbers of tanks equal to three.

To the plot is associated the variation of the ratio between the tank configuration mass and the propellant mass for each solution. As it is possible to observe, the mass tends to increase as the solutions move from the more vulnerable and more demisable solutions to the more survivable and less demisable solutions. However, there are solutions corresponding to configurations with a number of tanks lower than the maximum allowed, which have a lower mass thus producing the valleys in the graph. Usually spacecraft propellant tanks have a mass raging between 10% and 20% of the propellant mass stored, and they can reach 50% for high storage pressures [35]. In Figure 21, such solutions are highlighted in red in both plots. It is possible to observe that only aluminium alloy solutions belong to the typical mass range for the specific mission. However, some stainless steel solutions are not too far away from the considered range and, considering this is a preliminary design analysis, some of them could be included during the later stages of the mission design process. In addition, it is possible to observe that all three kinds of configuration (one, two, and three tanks) are present in the typical range, giving a wider variety of options to the design team.

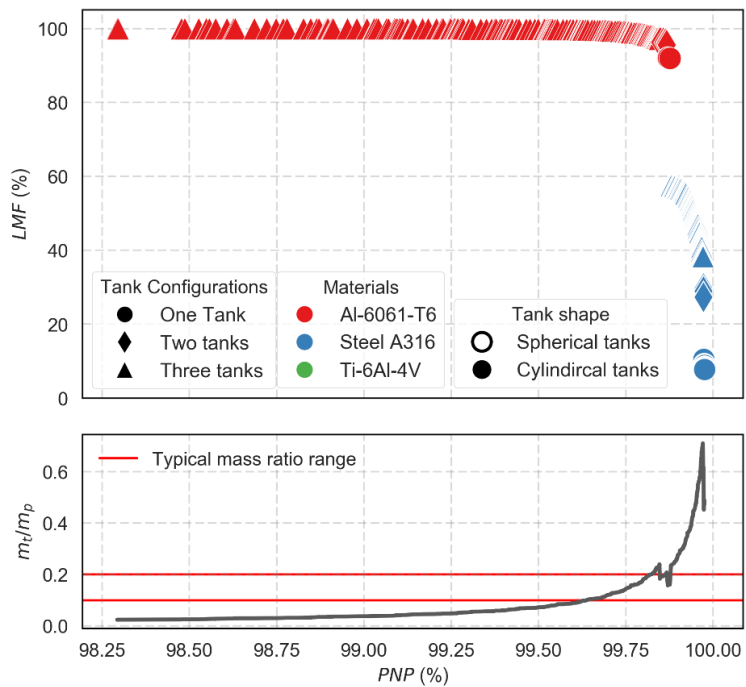

Figure 21: Pareto front for a 2000 kg, 7 years mission with maximum 3 tanks, with variation of the ratio between the tank configuration and the propellant mass. Highlighted (red lines and symbols) the solutions inside the typical range for actual missions.

Another important aspect to look at is the actual feasibility of the solutions, i.e. whether the tanks are actually able to sustain the normal operating pressures. Typically, a propellant tank has a storage pressure between 2 and 4 MPa. For cylindrical tanks, which constitute the solutions obtained in the Pareto front, the stress on the tanks' walls can be computed as follows:

$$\sigma_w = \frac{p r_t}{t_t} \qquad (51)$$

where $\sigma_w$ is the stress on the walls of the tank, and $p$ is the storage pressure. The computed tension has to be lower than the ultimate tensile strength of the material of the tank. This in turn imposes a limit on the maximum pressure that a tank with a certain radius, thickness and material can withstand, which is [35]:

$$p_{max} = \frac{\sigma_u \cdot t_t}{r_t \cdot SF} \qquad (52)$$

where $\sigma_u$ is the ultimate tensile strength of the material, and *SF* is a safety factor assumed equal to 1.5 for the current study. In Figure 22 the same Pareto front as before is presented. In this case the second plot represent the maximum sustainable storage pressure of the solutions belonging to the Pareto front and computed with Eq. (52). Even in this case, the solutions within the common limits (2-4 MPa) have been highlighted in red. As it is possible to observe, both aluminium alloy and stainless steel solutions belong to the highlighted range. Obviously, all the stainless steel solutions would be viable as they can resist a storage pressure higher than the 4 MPa limit. The aluminium alloy solutions outside the range, on the other hand, would be too thin to withstand the operating pressures normally used.

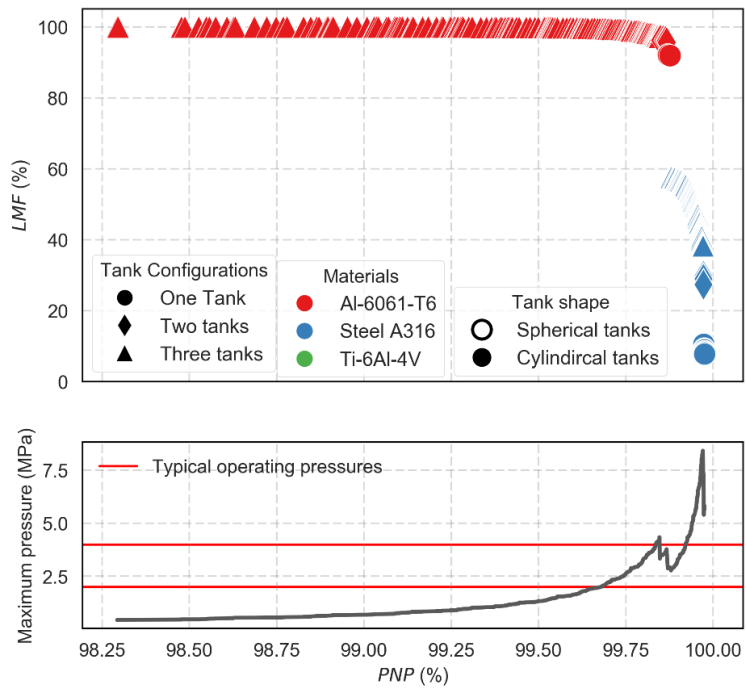

Figure 22: Pareto front for a 2000 kg, 7 years mission with maximum 3 tanks, with variation of the maximum storage pressure. Highlighted (red lines and symbols) the solutions inside the typical range of propellant storage pressures.

Finally, we can consider a more casualty risk related parameter that is the on-ground impact energy. As was mentioned, the demisability index adopted in this study has some limitations because it only considers the percentage of demised mass. However, it is also important to know the mass that actually lands on the surface, and even more important is to know the impact energy of the re-entering object. In fact, the NASA standard [3] states that a surviving object does not pose a risk for people on the ground if it has an impact energy below 15 J. This means the impact energy is an important parameter when considering the demisability of spacecraft components. In Figure 23, the impact energy of each solution in the Pareto front is represented.

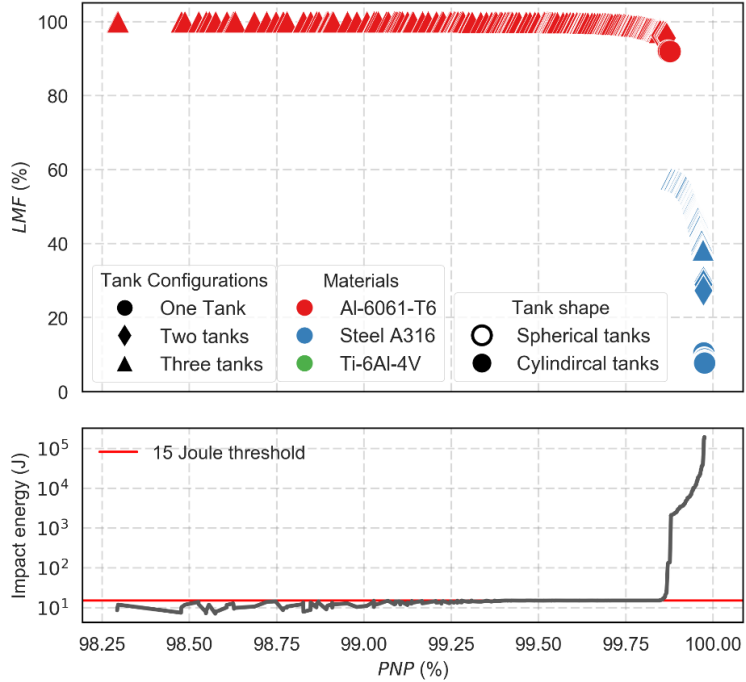

Figure 23: Pareto front for a 2000 kg, 7 years mission with maximum 3 tanks, with variation of the final energy of the re-entering object. Highlighted (red lines and symbols) the solutions below the 15 J threshold.

There are many solutions below the 15-joule threshold; however, all of them are made of aluminium alloy. Stainless steel solutions reaching the surface are usually too heavy to have such a low impact energy. However, the 15-joule limit is quite specific and, given the uncertainty intrinsic to re-entry simulations, it probably should not constitute a hard boundary to the validity of a solution, especially during a preliminary design phase. It is evident from this analysis of the solutions obtained from the optimisation that some improvements need to be introduced to the formulation of the optimisation itself. In fact, it is more efficient and more useful to obtain solutions that already take into account mission related constraints such as the overall mass of the system or its strength, as well as the casualty risk requirements. This can be obtained through a different definition of the fitness functions. For example, by changing the demisability index so that it includes a way to take into account the impact energy of the re-entering components. Moreover, a set of constraints can be introduced during the optimisation to take into account the limitations such as the maximum allowed storage pressure.

*5.2 Multi-objective optimisation discussion*

As introduced at the beginning of Section 4, the strategy proposed for the development of the multi-objective optimisation framework described in the paper is the one of evolutionary algorithms. The selection of these types of algorithms has been based on the characteristics they present with respect to the problem in exam. Given their population-based approach, genetic algorithms are well suited to solve multi-objective optimisation problems such as the one we are considering [55]. In addition, they allow the simultaneous search of different regions of the search space, making it possible to find a very diverse set of solutions for difficult problems were the search space is discontinuous. This is particularly true for the type of problem presented in the paper, where the search space is composed of both continuous and discrete variables. Moreover, the aim of the optimisation framework under development is the exploration of the large design space of preliminary spacecraft configuration in order to be able to find viable solutions that can be studied more in depth in later stages of the mission design process. This is exactly one of the situations where evolutionary algorithms excel. However, given their stochastic nature, evolutionary algorithms can only find an approximation of the Pareto front and not the actual front. Nonetheless, for the purpose of the developed framework, the aim is to find a set of preliminary solutions and the approximate nature of evolutionary algorithms can thus be accepted.

*5.2.1 Models and optimisation computational costs*

When looking at optimisation it is also important to look at the computational cost of the simulation. The developed code is entirely written in Python and the simulations for the article have been performed on an 8-core Intel i7 at 3.4 GHz PC with 16 GB of RAM. A summary of the average simulation times for a single simulation of both the demisability and survivability models are shown in Table 8, together with the average time for an optimisation simulation. The single simulations are not parallelised, whereas the optimisation cost is for a simulation parallelised on the 8 cores of the PC.

Table 8: Computational cost of the demisability model, survivability model, and optimisation simulation. The values in the table are the average over the different simulations performed in the study.

|  | Demisability model | Survivability model |
|---|---|---|
| Avg. single simulation time (s) | 2.46 | 1.6 |
| Avg. optimisation execution time (s) | 3557.2 | |

The simulation times for the models are not constant; they depend on the characteristics of the spacecraft configuration. For example, it is possible to observe from Figure 24 that the demisability model cost is influenced by both the thickness and size (number of vessels) of the tanks. This is a consequence of the different ballistic coefficients resulting for the different tanks configurations, which influence the shape of the trajectory. The survivability model (Figure 25) instead is more influenced by the number of vessels, as for each vessel the penetration probability as to be computed separately.

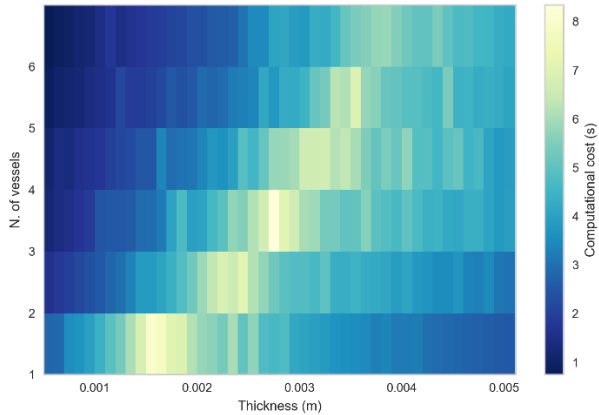 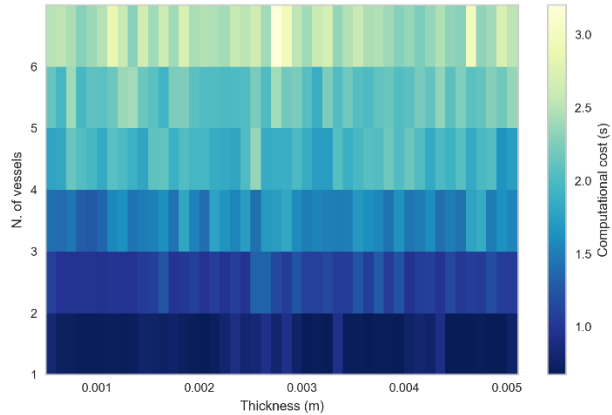

Figure 24: Demisability model computational cost for cylindrical aluminium alloy tanks as a function of the thickness and number of vessels.

Figure 25: Survivability model computational cost for cylindrical aluminium alloy tanks as a function of the thickness and number of vessels.

*5.2.2 Indices behaviour*

A further justification supporting the use of an evolutionary algorithm to explore the search space under the requirements of the demisability and of the survivability is given by the behaviour of the two indices as a function of the optimisation variables. An example of the discontinuous and nonlinear behaviour of the indices as a function of the tank thickness and material type is given in Figure 26 and Figure 27. The example refers to a two vessels cylindrical tank configuration. It is possible to observe how both indices have different values when changing the material, especially the demisability index. In addition, even the behaviour of the indices with respect to the thickness is not linear and can still have discontinuity. In the case of the demisability index, it follows an almost logarithmic trend. On the other end, the demisability index shows a very different behaviour depending on the material. For the titanium alloy a constant value of zero, for the stainless steel a dogleg curve with a maximum around 1.5 mm, and for the aluminium alloy, an initial flat curve shows a discontinuity around 3.6 mm.

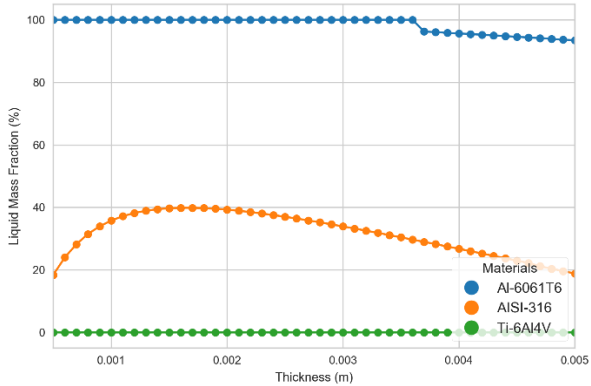

Figure 26: Demisability index trends as a function of the tank thickness and material for a two vessels assembly with cylindrical tanks.

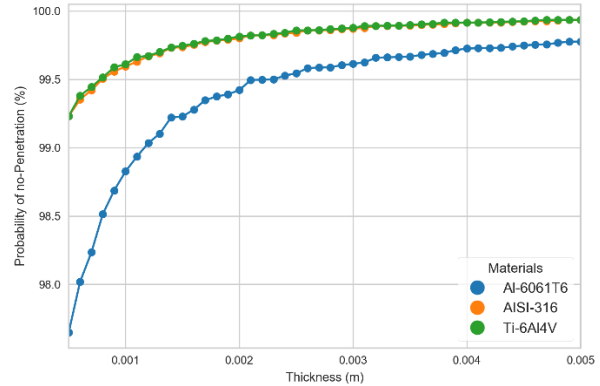

Figure 27: Survivability index trends as a function of the tank thickness and material for a two vessels assembly with cylindrical tanks.

*5.2.3 Future developments of the optimisation framework*

The optimisation framework has been applied to a simplified example, which was selected to illustrate the capabilities of the framework. It is meant to demonstrate the applicability and the efficacy of multi-objective optimisation using genetic algorithms to the wide exploration of preliminary spacecraft design solutions under the requirements of the demisability and of the survivability. The introduced optimisation framework is being developed in order to be able to analyse more complete satellite configurations, which include several components, of different types (e.g. tanks, reaction wheels, batteries, payloads, external panels, etc.). Not only the characteristics of the single components such as dimensions, thickness, material, and shape will be optimisable; also, their position inside the spacecraft and their possible attachment to the panels of the main structure will be considered. It will also be possible to handle component specific constraints and associate them with mission specific characteristics. For example, the minimum thickness required by a tank with given shape, material, size, and given the propellant required by the mission and the maximum specified storage pressure; or the minimum radius of a reaction wheel given its material, rotation speed, and mission required angular momentum.

Despite the dimensionality of the presented test case is low with only 4 optimisation variables, with the development of the framework, the complexity of the optimisation problem will quickly increase. For example, if we consider 4 optimisable components similar to the one used in the test case (i.e. 4 variables, 3 discrete and 1 continuous) and we add the possibility to optimise the position (3 continuous variables for each component), we pass from 4 to 28 variables (12 discrete and 16 continuous). We decided to use genetic algorithm with in mind the necessity to explore such large search spaces [56], characterised by both continuous and discrete variables. The example presented is used to demonstrate the applicability of the framework and shows the competing nature of the demisability and survivability requirements.

## 6 Conclusions

An optimisation framework that analyses preliminary spacecraft design options against the requirements posed by the demisability and the survivability has been proposed and described. The framework has then been applied to a simplified test case in order to demonstrate its characteristics and applicability, as well as the influence of the demisability and survivability requirements on the design choices. The test cases considered spacecraft in sun-synchronous missions and a specific type of internal component to optimise (tank assemblies). Two fitness functions describing the demisability and the survivability have been presented, and different tank configurations have been analysed as a function of their material, geometry, and configuration. The optimiser is able to provide a wide range of solutions for different types of mission scenarios. Some trends could be observed in the output of the optimisation for the different scenarios as a function of the mission class, lifetime and of the maximum allowed number of tanks in the configuration. For example, no titanium solutions were observed, nor were configurations with spherical tanks obtained.

A more in depth analysis of the solutions obtained showed that only a subset of the solutions belonged to typical configurations with respect to the mass and strength of the configuration. Moreover, only certain solutions in the Pareto front could satisfy the casualty risk criterion of the 15-joule threshold. This means that a further effort needs to be made in the definition of the optimisation problem through a more tailored definition of the demisability and survivability indices and through the application of constraints during the optimisation. Considering these additional properties may lead to a different set of solutions. For example, spherical tanks may be present because of their capability to withstand higher storage pressures with respect to cylindrical tanks with the same volume.

The developed optimisation framework has been introduced and tested on a simplified example. Future development of the framework will allow the optimisation of more complex spacecraft configurations, with many types of components (tanks, reaction wheels, batteries, etc.). In addition, the definition of mission and component specific constraints will be possible.

**Acknowledgments**

Part of this work was funded by EPSRC DTP/CDT through the grant EP/K503150/1. All data supporting this study are openly available from the University of Southampton repository at https://doi.org/10.5258/SOTON/D0036.Part of this work was funded by EPSRC DTP/CDT through the grant EP/K503150/1. All data supporting this study are openly available from the University of Southampton repository at https://doi.org/10.5258/SOTON/D0036.